\documentclass[journal]{IEEEtran}
\ifCLASSINFOpdf
\else
\fi
\usepackage{algorithm}
\usepackage{algpseudocode}
\usepackage{enumitem}
\usepackage{bm}
\usepackage{amssymb}
\usepackage{caption}
\usepackage{graphicx}
\usepackage{subfigure}
\usepackage{bbm}
\usepackage{colortbl}
\usepackage{mathtools}
\usepackage{makecell}
\usepackage{booktabs}
\usepackage{amsmath}
\usepackage{amsfonts}
\usepackage{color}
\usepackage{multirow}
\usepackage{booktabs}
\usepackage{xcolor}
\usepackage{arydshln}
\usepackage[switch]{lineno}

\usepackage[colorlinks,linkcolor=red]{hyperref}

\begin{document}
%
\title{Semantic Image Synthesis via Diffusion Models}
%
%
%

\author{Wengang Zhou,
        Weilun Wang,
        Jianmin Bao,
        Dongdong Chen,
        Dong Chen, 
        Lu Yuan, \\
        and~Houqiang Li,~\IEEEmembership{IEEE~Fellow}
\thanks{Corresponding author: Jianmin Bao}
\thanks{Wengang~Zhou, Weilun~Wang and Houqiang~Li are with the Department of Electrical Engineering and Information Science, 
        University of Science and Technology of China, Hefei, 230027 China 
        (e-mail: zhwg@ustc.edu.cn, wwlustc@mail.ustc.edu.cn, lihq@ustc.edu.cn).}
\thanks{Jianmin Bao, Dong Chen are with the Microsoft Research Asia, Beijing 100080, China. (e-mail: jianbao@microsoft.com, doch@microsoft.com).}
\thanks{Dongdong Chen, Lu Yuan are with the Microsoft Cloud+AI. (e-mail: dochen@microsoft.com, luyuan@microsoft.com).}}

%
%

\markboth{IEEE Transactions on Multimedia,~Vol.~**, No.~**, April~2025}%
{Shell \MakeLowercase{\textit{et al.}}: Bare Demo of IEEEtran.cls for IEEE Journals}
%



\maketitle

\begin{abstract}
Denoising Diffusion Probabilistic Models (DDPMs) have achieved remarkable success in various image generation tasks compared with Generative Adversarial Nets (GANs). 
Recent work on semantic image synthesis mainly follows the \emph{de facto} GAN-based approaches, which may lead to unsatisfactory quality or diversity of generated images.
In this paper, we propose a novel framework based on DDPM for semantic image synthesis.
Unlike previous conditional diffusion model directly feeds the semantic layout and noisy image as input to a U-Net structure, which may not fully leverage the information in the input semantic mask,
our framework processes semantic layout and noisy image differently.
It feeds noisy image to the encoder of the U-Net structure while the semantic layout to the decoder by multi-layer spatially-adaptive normalization operators. 
To further improve the generation quality and semantic interpretability in semantic image synthesis, we introduce the classifier-free guidance sampling strategy, which acknowledge the scores of an unconditional model for sampling process.
Extensive experiments on four benchmark datasets demonstrate the effectiveness of our proposed method, achieving state-of-the-art performance in terms of fidelity~(FID) and diversity~(LPIPS).
Our code and pretrained models are available at \url{https://github.com/WeilunWang/semantic-diffusion-model}.
\end{abstract}

\begin{IEEEkeywords}
Diffusion denoising probabilistic models, Semantic image synthesis, Image-to-image translation, Image generation.
\end{IEEEkeywords}

\section{Introduction}
Semantic image synthesis aims to generate photo-realistic images based on semantic layouts, which is a reverse problem of semantic segmentation.
This problem can be widely used in various applications, \emph{i.e.}, image editing, interactive painting and content generation.
Recent work~\cite{liu2019learning,TaesungPark2019SemanticIS,VadimSushko2020YouON,tang2020dual,tang2020local,wang2018high} mainly follows the adversarial learning paradigm, where the network is trained with adversarial loss~\cite{goodfellow2014generative}, along with a reconstruction loss.
By exploring the model architectures, these methods gradually improve performance on the benchmark datasets.
However, existing GAN-based approaches show limitations on some complex scenes in terms of generating high-fidelity and diverse results.

\begin{figure}[t]
    \centering
    \includegraphics[width=\linewidth]{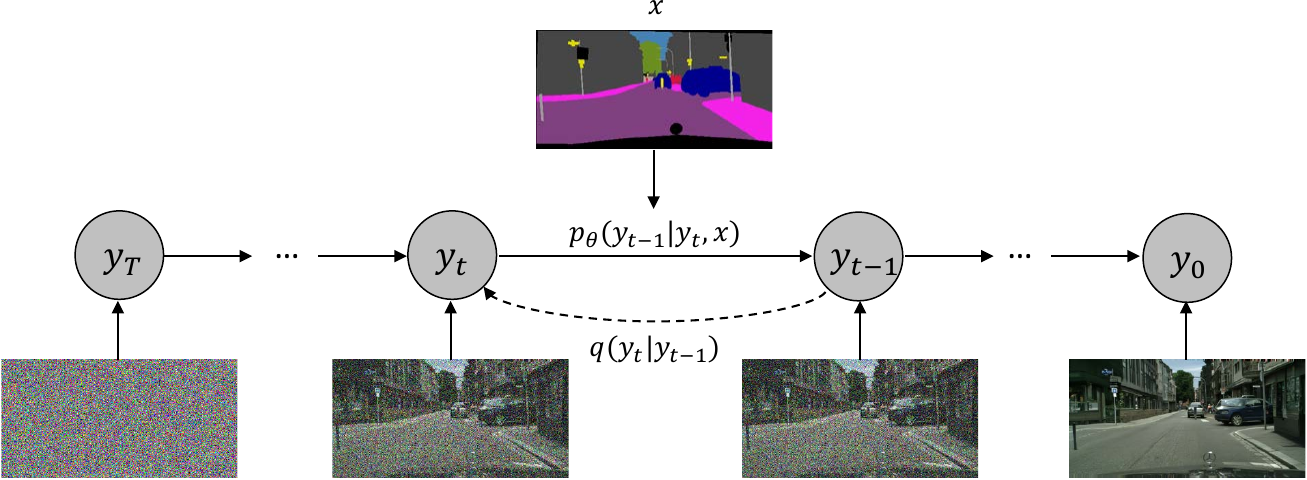}
    \caption{\textbf{Conditional Diffusion Model for Semantic Image Synthesis.}
    The framework transforms the noise from standard Gaussian distribution to the realistic image through iterative denoising process.
    In each denoising step, we use a U-net-based network to predict noise involved into the noisy images $y_t$ under the guidance of the semantic layouts $x$.
    }
    \vspace{-4mm}
    \label{fig:ddpm}
\end{figure}

Denoising diffusion probabilistic models (DDPMs)~\cite{ho2020denoising} is a new class of generative model based on maximum likelihood learning. DDPMs generate samples from standard Gaussian distribution to samples of an empirical distribution by an iterative denoising process. With the help of progressive refinement of the generated results, they achieve state-of-the-art sample quality on a number of image generation benchmarks~\cite{ho2020denoising, dhariwal2021diffusion,gu2021vector}.

In this paper, we present explore diffusion model for the problem of semantic image synthesis and design a novel framework named Semantic Diffusion Model (SDM). The framework follows the denoising diffusion paradigm, transforming the sampled Gaussian noise into a realistic image through an iterative denoising process (see Figure~\ref{fig:ddpm}). The generation process is a parameterized Markov chain. In each step, the noise is estimated from the input noisy image by a denoising network conditioned on the semantic label map. According to the estimated noise, a less noisy image is generated by the posterior probability formulation. Through iteration, the denoising network progressively produces semantic-related content and injects it into the stream to generate realistic images.

We revisit the previous conditional DDPMs~\cite{saharia2021palette,saharia2021image} that directly concatenate the condition information with the noisy image as input of the denoising network. The approach does not fully leverage the information in the input semantic mask, which leads to generated images in low quality and semantic relevance as suggested in previous work~\cite{TaesungPark2019SemanticIS}. Motivated by this, we design a conditional denoising network which processes semantic layout and noisy image independently.
The noisy image is fed into the encoder of the denoising network while the semantic layout is embedded into the the decoder of the denoising network by multi-layer spatially-adaptive normalization operators. This highly improves the quality and semantic correlation of generated images.

Furthermore, diffusion model are inherently capable of generating diverse results. The sampling strategy plays an important role in balancing quality and diversity of the generated results. The  na\"ive sampling procedure can generate images that demonstrate high diversity but lack the realism and strong correspondence with semantic label maps. Inspired by~\cite{ho2021classifier}, we adopt the classifier-free guidance strategy to boost image fidelity and semantic correspondence. Specifically, we fine-tune the pre-trained diffusion model by randomly removing the semantic mask input. Then the sampling strategy is processed based on both the predictions from diffusion model with and without semantic mask. By interpolating the scores from these two situations, the sampling results achieve a higher fidelity and stronger correlation with the semantic mask input.

To demonstrate the superiority of our framework, we conduct experiments on four benchmark datasets, \emph{i.e.}, Cityscapes, ADE20K, CelebAMask-HQ and COCO-Stuff. 
Both quantitative and qualitative results validates that our framework can generate both  high-fidelity and diverse results, achieving superior performance compared with previous methods.
Overall, the contributions are summarized as follows:
\begin{itemize}
    \item We propose a novel framework called \textbf{Semantic Diffusion Model} based on DDPMs, for high-fidelity and diverse semantic image synthesis.
    \item We find the network structure of current conditional diffusion models show limitation in handling the noisy input and semantic masks. We propose a new structure to handle noisy input and semantic mask separately and precisely. 
    \item To achieve better sampling results in diffusion process, we introduce the classifier-free guidance, which yields significantly higher quality and semantic input correlated results.
    \item  Extensive experiments on four benchmark datasets demonstrate the effectiveness of the proposed framework, achieving new state-of-the-art performance on generation fidelity (FID) and diversity (LPIPS).
\end{itemize}

\section{Related Work}
In this section, we briefly review the related topics, including denoising diffusion probabilistic models, image-to-image translation and semantic image synthesis.

\subsection{Denoising diffusion probabilistic models.}
A diffusion probabilistic model~\cite{sohl2015deep} is a parameterized Markov chain that optimizes the lower variational bound on the likelihood function to generate samples matching the data distribution.
The diffusion probabilistic model is efficient to define and train but is incapable of generating high-quality samples before.
Ho~\emph{et. al.}~\cite{ho2020denoising} first combine the diffusion probabilistic model with the score-based model and propose the denoising diffusion probabilistic model, which achieves great success in image generation. 
After that, more and more researchers~\cite{song2020denoising,ho2022cascaded,nichol2021improved} turn their attention to DDPMs.
Notably, Dhariwal and Nichol~\cite{dhariwal2021diffusion} show the potential of DDPMs, achieving image sample quality superior to GANs, on unconditional image generation.

Recently, conditional DDPMs~\cite{choi2021ilvr, avrahami2021blended, saharia2021image, saharia2021palette, meng2021sdedit} are studied to develop the application on downstream tasks.
Saharia~\emph{et. al.}~\cite{saharia2021image} achieve success in super resolution with DDPM.
Pattle~\cite{saharia2021palette} explores DDPM on four image-to-image translation problems, \emph{i.e.}, colorization, inpainting, uncropping, and JPEG decompression.
Bahjat~\emph{et. al.}~\cite{kawar2022denoising} propose an unsupervised posterior sampling method, \emph{i.e.}, DDRM, to solve any linear inverse problem with a pre-trained DDPM. Two concurrent works \cite{gu2021vector,nichol2021glide} apply DDPMs for text-to-image generation.
However, the aforementioned methods mainly focus on low-level computer vision tasks or work on single dimensional conditions.
Differently, we investigate conditional DDPM on generation problem with high-level dense semantic condition.

\begin{figure*}[t]
    \centering
    \includegraphics[width=\linewidth]{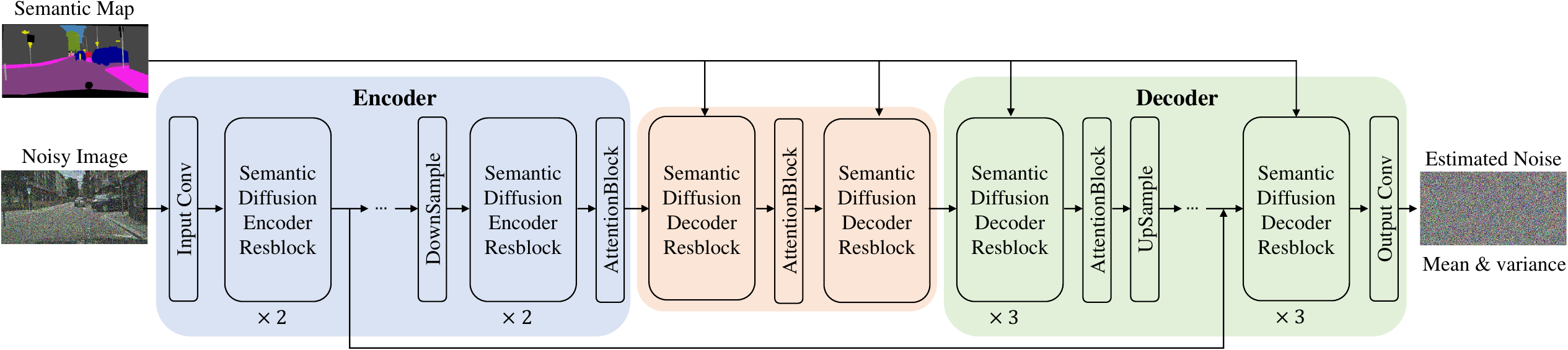}
    \vspace{-3mm}
    \caption{
    Overview of conditional denoising network, which mainly consists of semantic diffusion encoder resblocks, semantic diffusion decoder resblocks and attention blocks.
    The denoising network takes the noisy image as input and estimates the involved noise under the guidance of the semantic label map.
    }
    \vspace{-5mm}
    \label{fig:overview}
\end{figure*}

\subsection{Image-to-Image Translation.}
Image-to-image translation (I2I)~\cite{wang2018high,isola2017image,zhu2017toward,wu2019transgaga,royer2017xgan,liu2019few,DRIT,zhan2022bi,liu2023isfgan,huang2022unsupervised,zheng2023asynchronous} transfers images from source domain to target domain with the content information preserved.
Pix2Pix~\cite{isola2017image} first generalizes a class of conditional generation problems and formulates the Image-to-image translation problem.
Semantic image synthesis is also a kind of image-to-image translation problem, which translates images from the semantic label domain to the image domain.
Earlier methods on image-to-image translation~\cite{TaesungPark2019SemanticIS,wang2018high,isola2017image} are mainly based on CNN-structured networks and the adversarial training paradigm.
Pix2PixHD~\cite{wang2018high} develop a multi-scale framework for photographic image synthesis from pixelwise condition.
DRPAN~\cite{wang2020discriminative} decompose the image-to-image generation procedure into three iterated steps and gradually optimize the synthesized images on the local part.
However, these methods often synthesis images containing artifacts, and do not support diverse image generation.
IQ-VAE~\cite{zhan2022auto} involves transformer-based structure for more high-quality generation, designing an integrated quantization scheme and producing images in an auto-regressive manner.
MCL-Net~\cite{zhan2022marginal} performs diverse image translation by leveraging an additional exemplar image with self-correlation map.
Recently, a new technique, \emph{i.e.}, denoising diffusion probabilistic model, is capable of generating both realistic and diverse images, which may become a more suitable solution for image-to-image translation.

Some methods start to explore image-to-image translation problem based on the diffusion models.
SDEdit~\cite{meng2021sdedit} synthesizes the translation results by first adding noise to the input and iteratively denoising through a stochastic differential equation.
DDIBs~\cite{su2022dual} performs image-to-image translation relying on the inherent optimal transport properties of diffusion models.
These methods achieve success on several image-to-image translation problems, \emph{i.e.}, stroke-to-image translation and color conversion.
However, the aforementioned methods require an input in the form of RGB pixels, such as stroke painting or natural image, which are not suited for all the image-to-image translation problem, \emph{e.g.}, semantic image synthesis, whose input is a discrete label map.
Therefore, we propose a novel framework, \emph{i.e.}, SDM, to tackle semantic image synthesis.

\subsection{Semantic Image Synthesis.}
Semantic image synthesis~\cite{liu2019learning,TaesungPark2019SemanticIS,VadimSushko2020YouON,tang2020dual,tang2020local,wang2018high,zhu2020semantically,chen2017photographic,zhu2020sean,tan2021efficient,lv2022semantic} transforms semantic layouts into diverse realistic images.
Recent work on semantic image synthesis is GAN-based and trained with the adversarial loss along with the reconstruction loss.
Pix2PixHD~\cite{wang2018high} utilizes a multi-scale generator to produce high-resolution images from semantic label maps.
SPADE~\cite{TaesungPark2019SemanticIS} proposes spatially-adaptive normalization to better embed the semantic layouts into the generator. 
CLADE ~\cite{tan2021efficient} further improves the efficiency of SPADE by proposing a new class-adaptive normalization layer.
SCGAN~\cite{wang2021image} introduces a dynamic weighted network for semantic relevance, structure and detail synthesis.

The aforementioned methods mainly focus on generating real and semantically-corresponding unimodal result.
Paralleled with these methods, some other methods~\cite{zhu2017toward, yang2019diversity, zhu2020semantically, tan2021diverse} explore multimodal generation, which is also a core target for one-to-many problems like semantic image synthesis.
To tackle this issue, BicycleGAN~\cite{zhu2017toward} encourages bidirectional mapping between the generated image and latent code,
and DSCGAN~\cite{yang2019diversity} propose a simple regularization loss to penalize the generator from mode collapse.
More recently, INADE~\cite{tan2021diverse} proposes a framework that supports diverse generation at the instance level by instance-adaptive stochastic sampling.
However, these multimodal methods still fail to obtain satisfactory results on generation quality and learned correspondence.
It is non-trivial for existing GAN-based methods to achieve high generation fidelity and diversity at the same time.

Recently, some approaches~\cite{zhang2023adding, wei2023inferring, xue2023freestyle, wang2024semflow, lv2024place} have also attempted to explore semantic image generation based on denoising diffusion models.
ControlNet~\cite{zhang2023adding} introduces a trainable copy to add spatial conditioning controls to large, pretrained text-to-image diffusion models.
iPOSE\cite{wei2023inferring} proposes to infer parts from object shape and leverage it for improving semantic image synthesis.
FreestyleNet~\cite{xue2023freestyle} explores the freestyle capability of the large-scale text-to-image model.
Some of these approaches are not specifically designed for semantic images, and some relies on large-scale pre-trained text-to-image models.
To this end, we explore a new kind of approach to semantic image synthesis, \emph{i.e.}, conditional denoising diffusion probabilistic model, and achieve both better fidelity and diversity.

\vspace{-1mm}
\section{Methodology}
In this paper, we present a novel framework named Semantic Diffusion Model (SDM) based on DDPMs to transform semantic layouts into realistic images~(see Figure~\ref{fig:ddpm}).
With the iterative refinement, our framework generates high-quality images with fine-grained details.
The multimodal generation is also supported and the generation results exhibit high diversity, which benefits from the randomness continuously involved by noise at each step.
The rest of this section is organized as follows:
We begin with reviewing previous conditional denoising diffusion probabilistic models.
After that, we outline the architecture and objective functions of the semantic diffusion model. 
Finally, we present the classifier-free guidance adopted during inference.

\subsection{Preliminaries}
We first briefly review the theory of conditional denoising diffusion probabilistic models.
Conditional diffusion models aims to maximize the likelihood $p_\theta(\mathbf{y}_0|\mathbf{x})$ while the conditional data distribution follows $q(\mathbf{y}_0|\mathbf{x})$.
In conditional DDPM, two processes are defined, \emph{i.e.} the reverse process and the forward process.
The reverse process $p_\theta(\mathbf{y}_{0:T}| \mathbf{x})$ is defined as a Markov chain with learned Gaussian transitions beginning with $p(\mathbf{y}_{T}) \sim \mathcal{N}(0, \mathbf{I})$, which is formulated as follows,
\begin{equation}
    p_\theta(\mathbf{y}_{0:T}| \mathbf{x}) = p(\mathbf{y}_T)\prod_{t=1}^T p_\theta(\mathbf{y}_{t-1}| \mathbf{y}_t, \mathbf{x}),
\end{equation}
\begin{equation}
    p_\theta(\mathbf{y}_{t-1}| \mathbf{y}_t, \mathbf{x}) = \mathcal{N}(\mathbf{y}_{t-1};\bm{\mu}_\theta (\mathbf{y}_t, \mathbf{x}, t), \bm{\Sigma}_\theta (\mathbf{y}_t, \mathbf{x}, t)).
\end{equation}
The forward process $q(\mathbf{y}_{1:T}|\mathbf{y}_0)$ is defined as a process that progressively involves Gaussian noise into the data according to a variance schedule $\beta_1, \dots, \beta_T$, which is formulated as follows,
\begin{equation}
    q(\mathbf{y}_{t}|\mathbf{y}_{t-1}) = \mathcal{N}(\mathbf{y}_{t}; \sqrt{1-\beta_t}\mathbf{y}_{t-1}, \beta_t \mathbf{I}). \\
\end{equation}
With the notation $\alpha_t := \prod_{s=1}^t (1 - \beta_s)$, we have
\begin{equation}
    q(\mathbf{y}_{t}|\mathbf{y}_{0}) = \mathcal{N}(\mathbf{y}_{t}; \sqrt{\alpha_t}\mathbf{y}_{0}, (1 - \alpha_t) \mathbf{I}). \\
\end{equation}
The conditional DDPM is trained to optimize the upper variational bound on negative log likelihood.
Assuming $\bm{\Sigma}_\theta (\mathbf{y}_t, \mathbf{x}, t)$ as $\sigma_t \mathbf{I}$, the optimization target is equivalent to a denoising process as follows,
\begin{equation}
    \mathcal{L}_{t-1} = \mathbb{E}_{\mathbf{y}_0, \bm{\epsilon}} [\gamma_t \| \bm{\epsilon} - \bm{\epsilon}_\theta(\sqrt{\alpha_t} \mathbf{y}_0 + \sqrt{1 - \alpha_t} \bm{\epsilon}, \mathbf{x}, t) \|_2 ]
\end{equation}
where $\mathcal{L}_{t-1}$ is the loss function at the timestep $t-1$.
$\gamma_t$ is a constant about timestep $t$.

\subsection{Semantic Diffusion Model.}
Figure~\ref{fig:overview} gives an overview of the conditional denoising network in SDM, which is a U-Net-based network estimating the noise in the input noisy image.
Unlike previous conditional diffusion models, our denoising network processes the semantic label map and noisy image independently.
The noisy image is fed into the denoising network at the encoder part.
To fully leverage the semantic information, the semantic label map is injected into the the decoder of the denoising network by multi-layer spatially-adaptive normalization operators.

\begin{figure}[t]
    \centering
    \includegraphics[width=\linewidth]{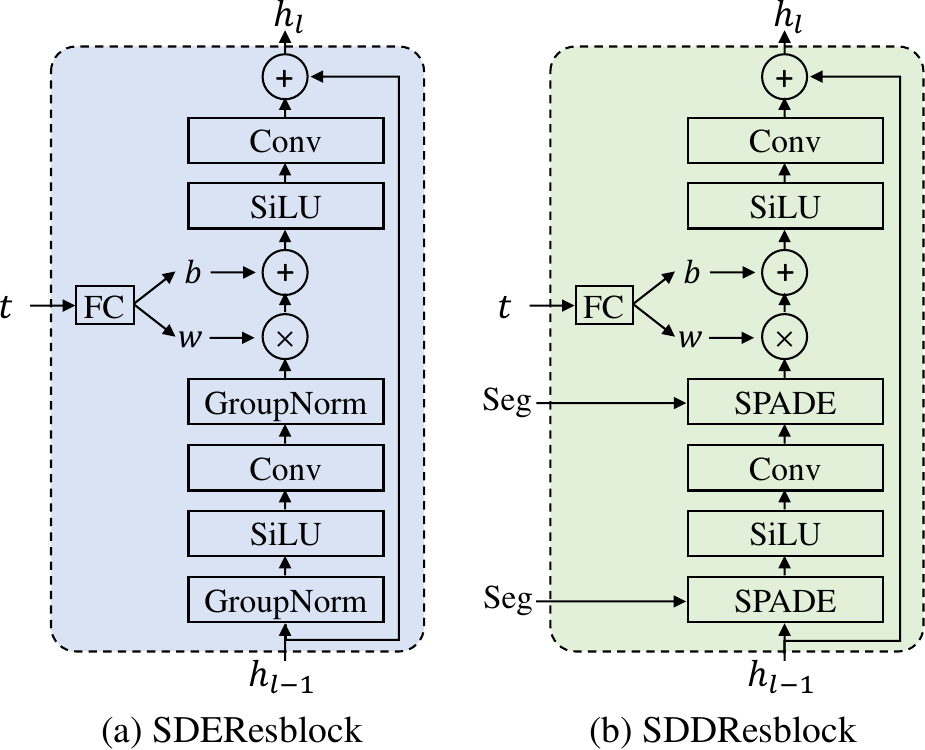}
    \caption{
    The detailed structure of semantic diffusion encoder resblock (SDEResblock) and semantic diffusion decoder resblock (SDDResblock).
    }
    \vspace{-4mm}
    \label{fig:SDResblock}
\end{figure}

\vspace{2mm}
\noindent \textbf{Encoder.}
We encode the feature of the noisy image with stacked semantic diffusion encoder resblocks (SDEResblocks) and attention blocks.
We show the detailed structure of the SDEResblocks in Figure~\ref{fig:SDResblock} (a).
To make the network estimate noise at different timestep $t$, SDEResblock involves $t$ by scaling and shifting the intermediate activation with learnable weight $\mathbf{w}(t) \in \mathbb{R}^{1 \times 1 \times C}$ and bias $\mathbf{b}(t) \in \mathbb{R}^{1 \times 1 \times C}$, which is formulated as follows,
\begin{equation}
    \mathbf{f}^{i+1} = \bm{w}(t) \cdot \mathbf{f}^{i} + \bm{b}(t),
\end{equation}
where $\mathbf{f}^{i}, \mathbf{f}^{i+1} \in \mathbb{R}^{H \times W \times C}$ are the input and output features, respectively.
The attention block refer to a self-attention layer~\cite{wang2018non} with skip connection, which is formulated as follows,
\begin{equation}
\begin{aligned}
    &\mathbf{f(x)} = \mathbf{W}_f \mathbf{x}, \enspace \mathbf{g(x)} = \mathbf{W}_g \mathbf{x}, \enspace \mathbf{h(x)} = \mathbf{W}_h \mathbf{x}, \\
    &\mathcal{M}(u, v) = \frac{\mathbf{f}(\mathbf{x}_u)^\top \mathbf{g}(\mathbf{x}_v)}{\|\mathbf{f}(\mathbf{x}_u)\|\|\mathbf{g}(\mathbf{x}_v)\|}, \\
    &\mathbf{y}_u = \mathbf{x}_u + \mathbf{W}_v \sum_v \text{softmax}_v(\alpha \mathcal{M}(u, v)) \cdot \mathbf{h}(\mathbf{x}_v),
\end{aligned}
\end{equation}
where $\mathbf{x}$ and $\mathbf{y}$ are the input and output of the attention block.
$\mathbf{W}_f$, $\mathbf{W}_g$, $\mathbf{W}_h$ and $\mathbf{W}_v$ $\in \mathbb{R}^{C \times C}$ refer to 1 $\times$ 1 convolution in the attention block, respectively.
$u$ and $v$ is the index of spatial dimension, range from 1 to $H \times W$.
We adopt the attention block on the feature at a specific resolution, \emph{i.e.}, 32 $\times$ 32, 16 $\times$ 16 and 8 $\times$ 8.

\vspace{2mm}
\noindent \textbf{Decoder.}
We inject the semantic label map into the decoder of the denoising network to guide the denoising procedure.
Revisiting the previous conditional diffusion models~\cite{saharia2021image, saharia2021palette} which directly concatenate the condition information with the noisy image as input, we find that this approach does not fully leverage the semantic information. Inspired by SPADE~\cite{TaesungPark2019SemanticIS}, one important reason is that the normalization layers in the denoising U-Net ``wash away'' the semantic information. More precisely, supposing that a segmentation mask with a single label is given as input, the convolution outputs are again uniform, with different labels having different uniform values, while the normalization outputs will become all zeros no matter what the input semantic label is given. This will finally lead to the generated images in low quality and weak semantic relevance.

\begin{figure*}[t]
    \centering
    \includegraphics[width=0.7\linewidth]{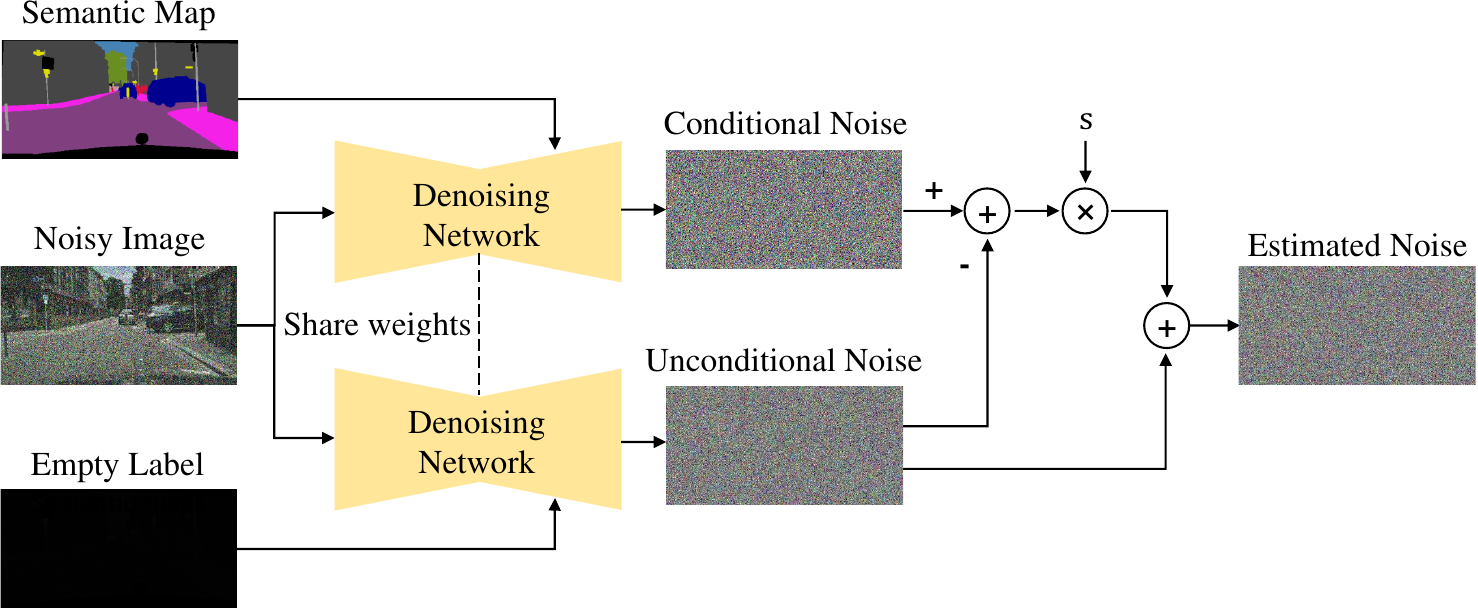}
    \caption{
    The sampling procedure with classifier-free guidance.
    }
    \vspace{-4mm}
    \label{fig:cfg}
\end{figure*}

To address this issue, we design the semantic diffusion decoder resblock (SDDResblock) (see Figure~\ref{fig:SDResblock} (b)) to embed the semantic label map into the the decoder of the denoising network in multi-layer spatially-adaptive manner.
Different from SDEResblock, we introduce the spatially-adaptive normalization (SPADE)~\cite{TaesungPark2019SemanticIS} instead of the group normalization.
The SPADE injects the semantic label map into the denoising streams by regulating the feature in a spatially-adaptive, learnable transformation, which is formulated as follows,
\begin{equation}
    \mathbf{f}^{i+1} = \bm{\gamma}^i(\mathbf{x}) \cdot \mathrm{Norm}(\mathbf{f}^i) + \bm{\beta}^i(\mathbf{x}),
\end{equation}
where $\mathbf{f}^{i}$ and $\mathbf{f}^{i+1}$ are the input and output features of SPADE.
$\mathrm{Norm}(\cdot)$ refers to the parameter-free group normalization.
$\bm{\gamma}^i(\mathbf{x})$, $\bm{\beta}^i(\mathbf{x})$ are the spatially-adaptive weight and bias learned from the semantic layout, respectively. It is worth mentioning that our framework is different from SPADE~\cite{TaesungPark2019SemanticIS}, since our SDM is specifically designed for diffusion process with attention block, skip-connection, and timestep embedding module while SPADE does not.

\begin{algorithm}[t] 
\caption{Finetuning Procedure.} 
\begin{algorithmic}[1]
\While {not converged} 
\State $\mathbf{x} \sim q(\mathbf{x}), \mathbf{y} \sim q(\mathbf{y}_0| \mathbf{x})$
\State $\bm{\epsilon} \sim \mathcal{N}(0, \mathbf{I}), t \sim \mathrm{uniform}\{1, 2, \dots, T\}$;
\State $\hat{\mathbf{x}} = \mathbf{x}~\text{if}~\mathrm{rand}() > 0.2,~\text{else}~\emptyset.$
\State $\widetilde{\mathbf{y}} = \sqrt{\alpha_t} \mathbf{y} + \sqrt{1 - \alpha_t} \bm{\epsilon}$;
\State Take a gradient descent step on
$$\nabla_\theta \mathcal{L}_\text{simple}(\bm{\epsilon}_\theta(\widetilde{\mathbf{y}}, \mathbf{\hat{x}}, t)) + \lambda \mathcal{L}_\text{vlb}(\bm{\epsilon}_\theta(\widetilde{\mathbf{y}}, \mathbf{\hat{x}}, t), \bm{\Sigma}_\theta(\widetilde{\mathbf{y}}, \mathbf{x}, t))$$
\EndWhile
\end{algorithmic}
\label{algo:train} 
\end{algorithm}

\begin{algorithm}[t] 
\caption{Inference Procedure in $T$ denoising steps.} 
\begin{algorithmic}[1] 
\State $y_T \sim \mathcal{N}(0, \mathbf{I})$;
\For{$t = T, \dots, 1$}
\State $\mathbf{z} \sim \mathcal{N}(0, \mathbf{I})~\text{if}~t \neq 0,~\text{else}~\mathbf{z} = \mathbf{0};$
\State $\hat{\epsilon}_\theta(y_t|x) = \epsilon_\theta(y_t|\emptyset) + s \cdot (\epsilon_\theta(y_t|x) - \epsilon_\theta(y_t|\emptyset));$
\State $y_{t-1} = \frac{1}{\sqrt{\alpha_t}}(y_t - \frac{1 - \alpha_t}{\sqrt{1-\beta_t}}\hat{\epsilon}_\theta(y_t|x) + \Sigma(y_t | x)^{\frac{1}{2}} \mathbf{z});$ 
\EndFor
\State \textbf{return} $y_0$
\end{algorithmic} 
\label{algo:inference} 
\end{algorithm}

\subsection{Classifier-free guidance.}
Following the common sampling procedure in DDPM, it is noticed that the generated images are diverse but not photo-realistic and not strongly correlated with the semantic label maps.
We hypothesis that the conditional diffusion model can not handle conditional input explicitly during the sampling process.
Previous method~\cite{dhariwal2021diffusion} discovered that samples from conditional diffusion models can often be improved by the gradient of the log probability $\nabla_{y_t} \text{log}~p(x|y_t)$. 
Assuming a conditional diffusion model with estimated mean $\mu_\theta(y_t|x)$ and variance $\Sigma_\theta(y_t|x)$, the results can be improved by perturbing the mean, which is formulated as follows,
\begin{equation}
    \hat{\mu}_\theta(y_t|x) = \mu_\theta(y_t|x) + s \cdot \Sigma_\theta(y_t|x) \cdot \nabla_{y_t} \text{log}~p(x|y_t)
\end{equation}
where the hyper-parameter $s$ is named the guidance scale, which trades off the sample quality and diversity.

\begin{table*}[t]
    \footnotesize
    \centering
    \caption{\textbf{Quantitative comparison with existing methods on semantic image synthesis.}
    $\uparrow$ indicates the higher the better, while $\downarrow$ indicates the lower the better.
    }
    \vspace{-2mm}
    \begin{tabular}{l cc cc cc cc}
    \toprule
    \multirow{2}{*}{\textbf{Method}} & \multicolumn{2}{c}{\textbf{CelebAMask-HQ}} & \multicolumn{2}{c}{\textbf{Cityscapes}} & \multicolumn{2}{c}{\textbf{ADE20K}} & \multicolumn{2}{c}{\textbf{COCO-Stuff}} \\
    \cmidrule(r){2-3} \cmidrule(r){4-5} \cmidrule(r){6-7} \cmidrule(r){8-9} 
    & {\bf FID}$\downarrow$ & {\bf LPIPS}$\uparrow$ & {\bf FID}$\downarrow$ & {\bf LPIPS}$\uparrow$ & {\bf FID}$\downarrow$ & {\bf LPIPS}$\uparrow$ & {\bf FID}$\downarrow$ & {\bf LPIPS}$\uparrow$ \\
    \midrule
    {Pix2PixHD~\cite{wang2018high}} & 38.5 & 0 & 95.0 & 0 & 81.8 & 0 & 111.5 & 0 \\
    {SPADE~\cite{TaesungPark2019SemanticIS}} & 29.2 & 0 & 71.8 & 0 & 22.6 & 0  & 33.9 & 0 \\
    {DAGAN~\cite{tang2020dual}} & 29.1 & 0 & 60.3 & 0 & 31.9 & 0  & n/a & 0 \\
    {SCGAN~\cite{wang2021image}} & 20.8 & 0 & 49.5 & 0 & 29.3 & 0 & 18.1 & 0 \\ 
    {CLADE~\cite{tan2021efficient}} & 30.6 & 0 & 57.2 & 0 & 35.4 & 0  & 29.2 & 0 \\ \cdashline{1-9}
    {CC-FPSE~\cite{liu2019learning}} & n/a & n/a & 54.3 & 0.026 & 31.7 & 0.078  & 19.2 & 0.098  \\ 
    {GroupDNet~\cite{zhu2020semantically}} & 25.9 & 0.365 & 47.3 & 0.101 & 41.7 & 0.230  & n/a & n/a \\
    {INADE~\cite{tan2021diverse}} & 21.5 & 0.415 & 44.3 & 0.295 & 35.2 & 0.459  & n/a & n/a \\
    {OASIS~\cite{VadimSushko2020YouON}} & n/a & n/a & 47.7 & 0.327 & 28.3 & 0.286 & 17.0 & 0.328 \\ 
    {ControlNet~\cite{zhang2023adding}} & 24.0 & 0.528 & 43.5 & 0.527 & 29.9 & 0.646 & 36.6 & 0.671 \\ 
    \cdashline{1-9}
    {SDM-LoRA~(Ours)} & 34.6 & \textbf{0.506} & 47.2 & \textbf{0.474} & 38.0 & \textbf{0.619} & 33.9 & \textbf{0.647} \\
    {SDM~(Ours)} & \textbf{18.8} & 0.422 & \textbf{42.1} & 0.362 & \textbf{27.5} & 0.524 & \textbf{15.9} & 0.518 \\
    \bottomrule
    \end{tabular}
    \label{tab:comparison}
\end{table*}

\begin{table*}[t]
    \footnotesize
    \centering
    \caption{Paired user study on four benchmark datasets between our method and several challenging methods, \emph{i.e.,} SPADE~\cite{TaesungPark2019SemanticIS}, INADE~\cite{tan2021diverse} and OASIS~\cite{VadimSushko2020YouON}.
    The reported numbers refer to the percentage of user preferences in favor of our approach.
    It is observed that our method is clearly preferred over the competitors on four benchmark datasets.
    }
    \vspace{-2mm}
    \begin{tabular}{l @{\hskip 15mm} c @{\hskip 15mm} c @{\hskip 15mm} c @{\hskip 15mm} c}
    \toprule
    \textbf{Method}  & \textbf{Cityscapes} & \textbf{ADE20K} & \textbf{CelebAMask-HQ} & \textbf{COCO-Stuff}\\ \midrule
    {SDM v.s. SPADE} & 84.0\% & 87.5\% & 76.5\% & 94.0\% \\
    {SDM v.s. INADE} & 75.5\% & 93.5\% & 89.0\% & n/a \\
    {SDM v.s. OASIS} & 84.0\% & 80.0\% & n/a & 84.0\% \\
    \bottomrule
    \end{tabular}
    \label{tab:user_study}
\end{table*}

Previous work~\cite{dhariwal2021diffusion} applied an extra trained classifier $p_\phi(x|y_t)$ to provide the gradient during sampling process. Inspired by~\cite{ho2021classifier}, we obtain from the guidance with the generative model itself instead of a classifier model that requires extra cost for training.
The main idea is to replace the semantic label map $x$ with a null label $\emptyset$ to disentangle the noise estimated under the guidance of semantic label map $\epsilon_\theta(y_t|x)$ from unconditional situation $\epsilon_\theta(y_t|\emptyset)$, as shown in Algorithm~\ref{algo:train}. 
The disentangled component implicitly infers the gradient of the log probability, which is formulated as follows,
\begin{equation}
\begin{aligned}
    \epsilon_\theta(y_t|x) - \epsilon_\theta(y_t|\emptyset) & \propto \nabla_{y_t} \text{log}~p(y_t|x) - \nabla_{y_t} \text{log}~p(y_t) \\
    & \propto \nabla_{y_t} \text{log}~p(x|y_t).
\end{aligned}
\label{eq:cfg}
\end{equation}
During sampling procedure, the disentangled component is increased to improve the samples from conditional diffusion models, which is formulated as follows,
\begin{equation}
    \hat{\epsilon}_\theta(y_t|x) = \epsilon_\theta(y_t|x) + s \cdot (\epsilon_\theta(y_t|x) - \epsilon_\theta(y_t|\emptyset)).
\end{equation}
In our implementation, $\emptyset$ is defined as the all-zero vector. We show the detailed sampling procedure in Figure~\ref{fig:cfg} and Algorithm~\ref{algo:inference}.

\section{Experiments}
\subsection{Experimental Setup}
\noindent \textbf{Datasets.}
We conduct experiments on four benchmark datasets, \emph{i.e.}, Cityscapes~\cite{cordts2016cityscapes}, ADE20K~\cite{zhou2017scene}, CelebAMask-HQ~\cite{CelebAMask-HQ} and COCO-Stuff~\cite{caesar2018coco}.
For the Cityscapes dataset, we apply one-hot activation of 35 classes as the input semantic label map.
Furthermore, inspired by SPADE [31], we produce an instance edge map from provided instance labels and concatenate it with a semantic label map as additional condition information.
The CelebAMask-HQ dataset is processed similarly to the Cityscapes dataset, taking one-hot activation of 19 classes and instance edge map as input.
For the ADE20K dataset, we apply one-hot activation of 151 classes (including an ``unknown'' object) as the input semantic label map.
The instance edge map is not employed on ADE20K dataset since the instance labels are not available.
On the COCO-Stuff dataset, we utilize one-hot activation of 183 classes (including an ``unknown'' class) and the instance labels.
For Cityscapes dataset, we resize images to the resolution of 256 $\times$ 512 for training.
For ADE20K, CelebAMask-HQ and COCO-Stuff dataset, we train our network on the resolution of 256 $\times$ 256.

\begin{figure*}[t]
    \centering
    \includegraphics[width=\linewidth]{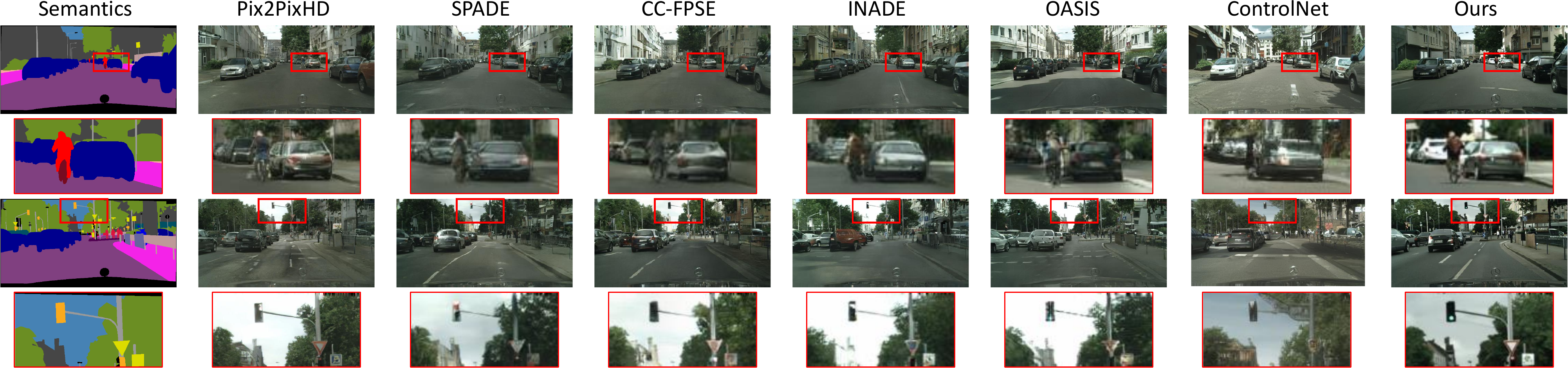}
    \caption{\textbf{Qualitative results on Cityscapes datasets.}
    We compare our method with several challenging methods, \emph{i.e.}, Pix2PixHD~\cite{wang2018high}, SPADE~\cite{TaesungPark2019SemanticIS}, CC-FPSE~\cite{liu2019learning}, INADE~\cite{tan2021diverse}, OASIS~\cite{VadimSushko2020YouON} and ControlNet~\cite{zhang2023adding}. 
    We present zoomed-in results of the generated images.
    Our method generates more reasonable and distinct results on fine-grained objects, such as distant cars and traffic lights.
    }
    \vspace{-5mm}
    \label{fig:comparison_city}
\end{figure*}

\begin{figure}[t]
    \centering
    \includegraphics[width=\linewidth]{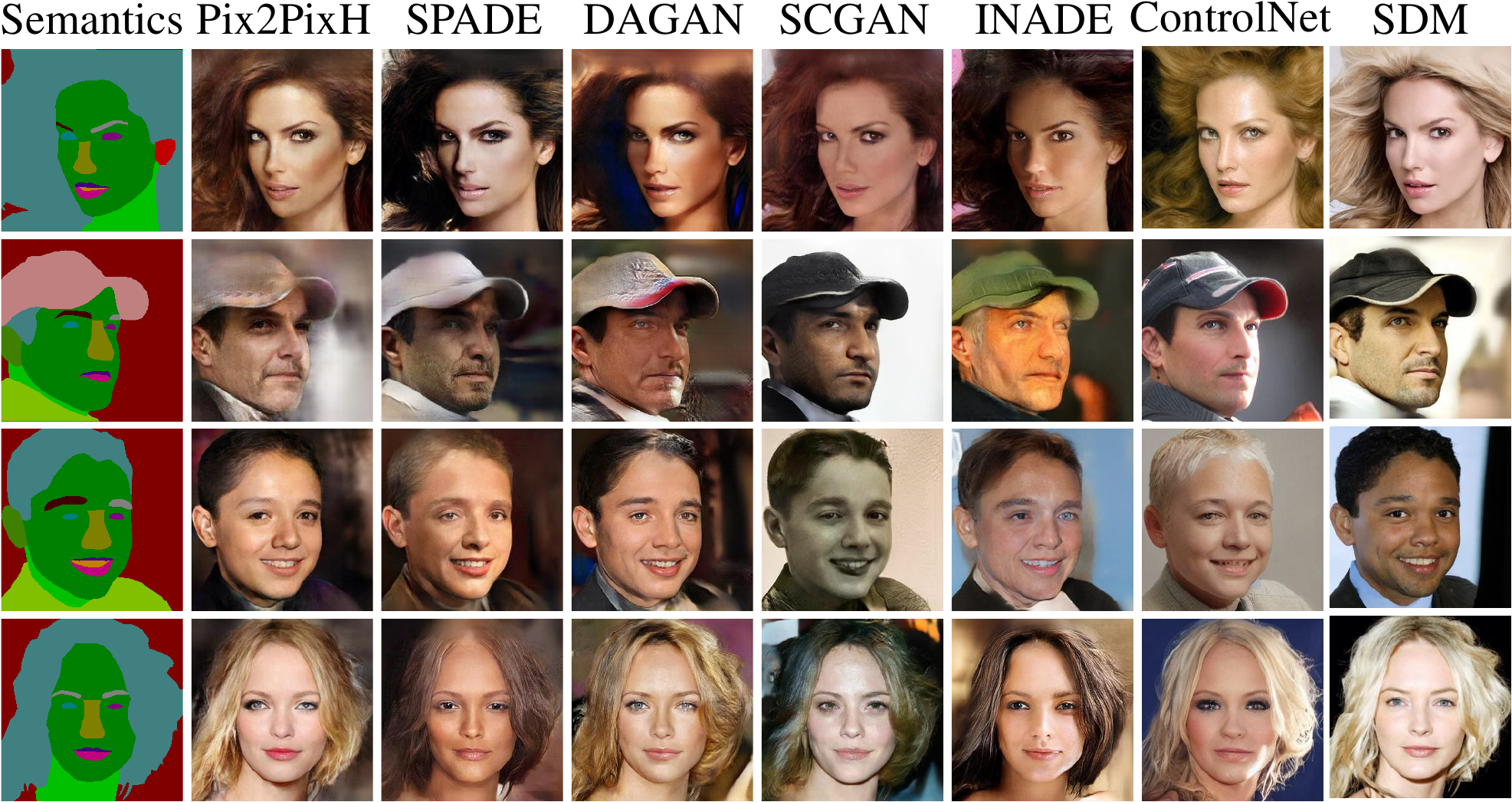}
    \caption{\textbf{Qualitative results on the CelebAMask-HQ dataset.}
    We compare our SDM method with several challenging methods, \emph{i.e.}, Pix2PixHD~\cite{wang2018high}, SPADE~\cite{TaesungPark2019SemanticIS}, DAGAN~\cite{tang2020dual}, SCGAN~\cite{wang2021image}, INADE~\cite{tan2021diverse} and ControlNet~\cite{zhang2023adding}.
    By comparison, our generated images show superior performance on fidelity and learned correspondence, especially on side face.
    }
    \label{fig:comparison_celeba}
    \vspace{-4mm}
\end{figure}

\begin{figure}[t]
    \centering
    \includegraphics[width=\linewidth]{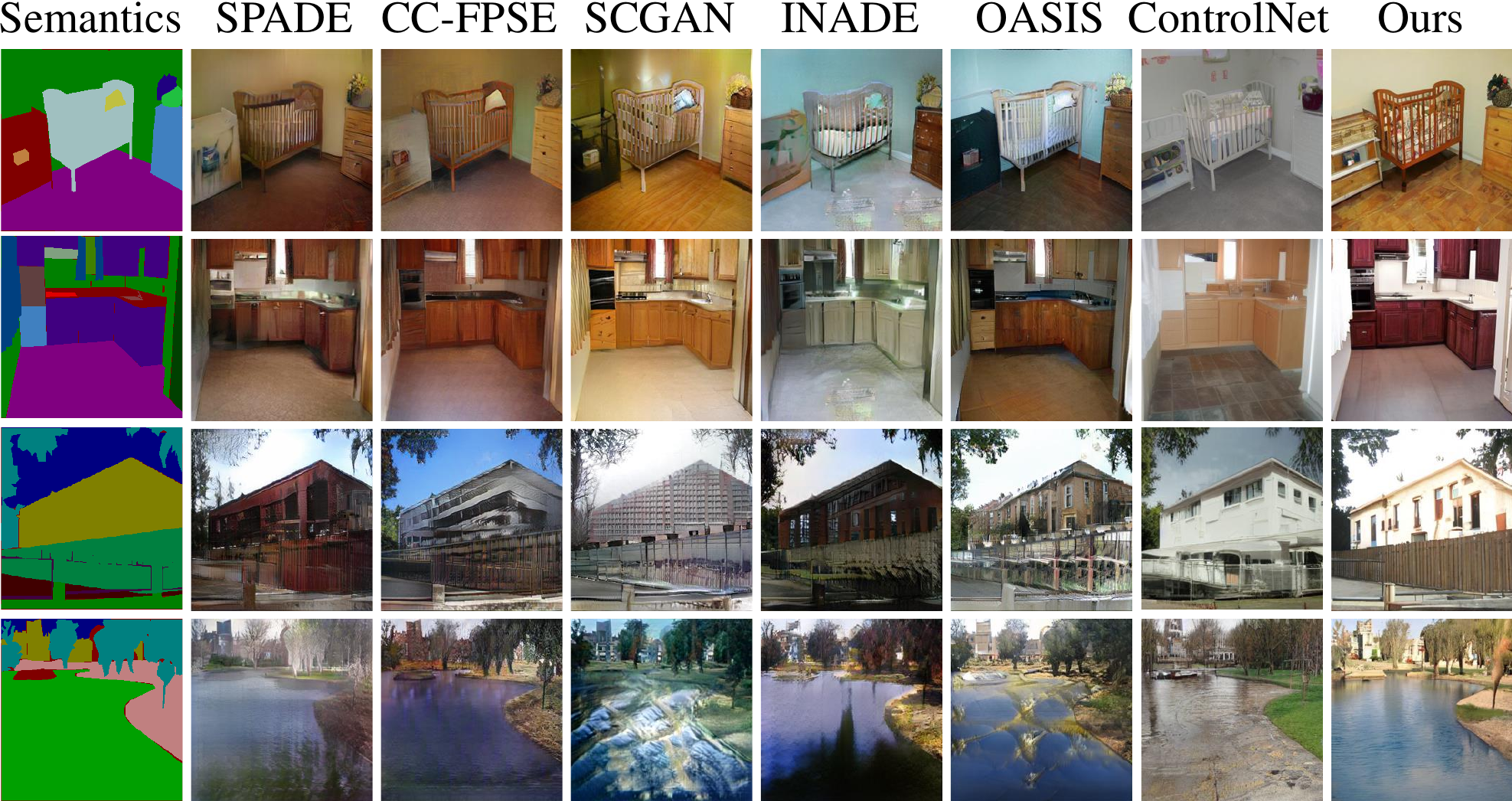}
    \caption{\textbf{Qualitative results on the ADE20K dataset.}
    We compare with the several challenging methods, \emph{i.e.}, SPADE~\cite{TaesungPark2019SemanticIS}, CC-FPSE~\cite{liu2019learning}, SCGAN~\cite{wang2021image}, INADE~\cite{tan2021diverse}, OASIS~\cite{VadimSushko2020YouON} and ControlNet~\cite{zhang2023adding}.
    By comparison, our method shows more reasonable generation results and better exhibits fine-grained details, \emph{i.e.}, water surface and fence.
    }
    \vspace{-4mm}
    \label{fig:comparison_ade}
\end{figure}

\noindent \textbf{Implementation details.}
Following DDPM [11], we set the total diffusion timestep to 1000.
In the forward process, the Gaussian noise is involved in the data according to a variance schedule $\beta_1, \dots, \beta_T$.
In our implementation, the variance schedule is arranged linearly with respect to the timestep $t$.
During the sampling procedure, we utilize the classifier-free guidance strategy.
The classifier-free guidance perturbs the mean as Equation~\ref{eq:cfg}.
In addition to the mean value, the denoising network also estimates the variance at timestep $t$, $\Sigma_\theta(\widetilde{y}, x, t)$.
The variance $\Sigma_\theta(\widetilde{y}, x, t)$ is not perturbed in classifier-free guidance.

The hyperparameters in the framework are set as follows:
Following the \cite{nichol2021improved}, we set the trade-off parameter $\lambda$ as 0.001 to ensure the training stability.
Since different datasets have different complexity, we apply different guidance scales $s$ on four datasets.
Guidance scale $s$ is set to 1.5, 2.0, 1.5 and 1.5 on the CelebAMask-HQ, Cityscapes, ADE20K and COCO-Stuff dataset, respectively.
We utilize AdamW optimizer~\cite{IlyaLoshchilov2018DecoupledWD} to train the framework.
During training, we adopt an exponential moving average (EMA) of the denoising network weights with 0.9999 decay.
The framework is implemented by Pytorch and experiments are performed on NVIDIA Tesla V100.

\begin{figure}[t]
    \centering
    \includegraphics[width=\linewidth]{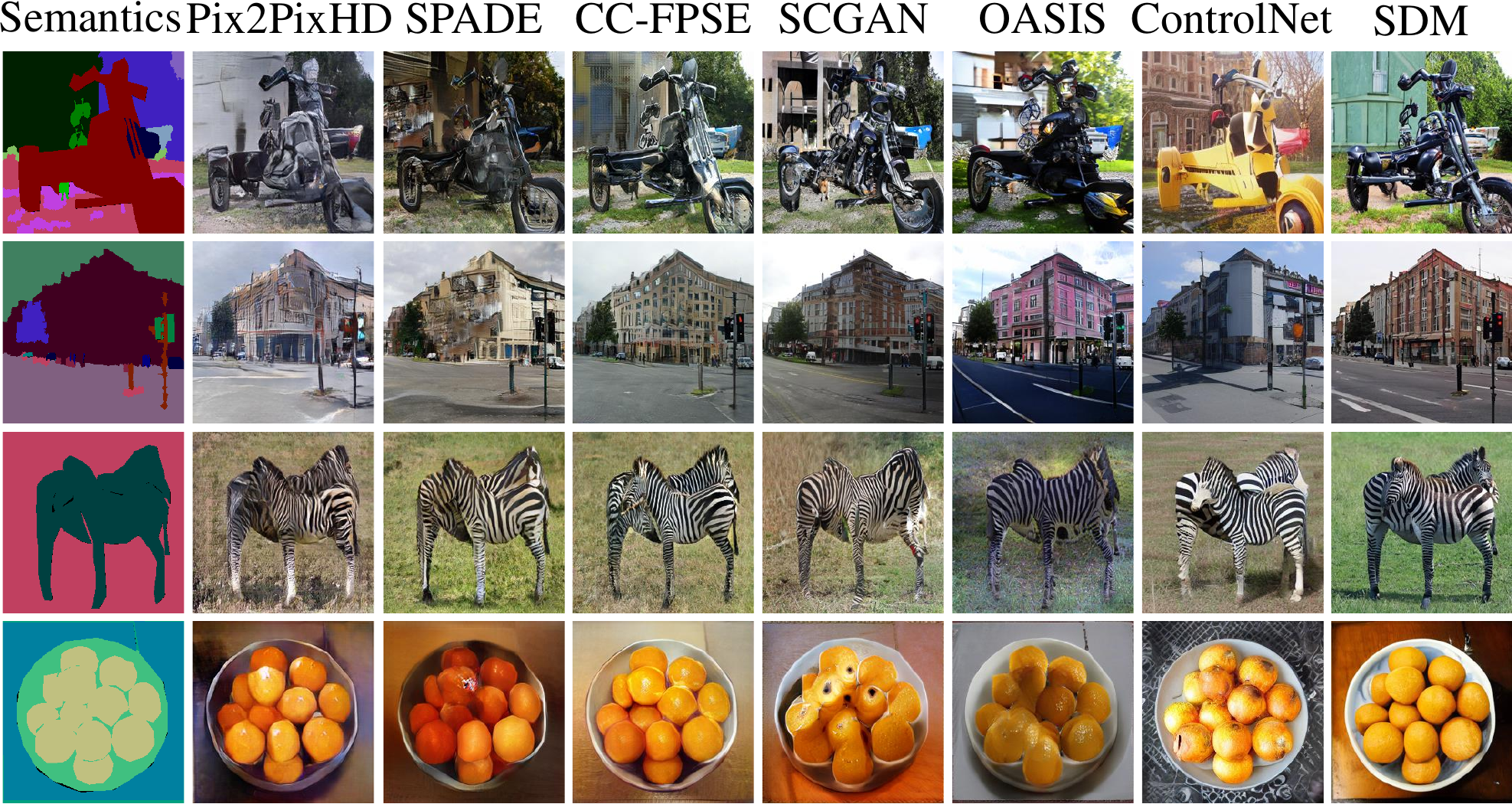}
    \caption{Qualitative results on the COCO-Stuff dataset.
    We compared our method with the several challenging methods, \emph{i.e.}, Pix2PixHD~\cite{wang2018high}, SPADE~\cite{TaesungPark2019SemanticIS}, CC-FPSE~\cite{liu2019learning}, SCGAN~\cite{wang2021image}, OASIS~\cite{VadimSushko2020YouON} and ControlNet~\cite{zhang2023adding}.
    By comparison, our method can better generate objects with complex structure. 
    }
    \vspace{-4mm}
    \label{fig:comparison_coco}
\end{figure}

\noindent \textbf{Evaluation.} 
We aim to assess visual quality, diversity and learned correspondence of generated images.
For the visual quality, we adopt the widely-used Fréchet Inception Distance~(FID) metrics.
To evaluate the generation diversity of different methods, we compute the average distance measured by the LPIPS metrics~\cite{zhang2018unreasonable} between multimodal generation results. For qualitative comparison, we try to compare all the methods but find some models are not publicly available and we also tried to email the author. We then choose the most recent and representative methods whose models are available for testing.

For the learned correspondence, we utilize an off-the-shelf network to evaluate the ``semantic interpretability'' of generated results.
We use DRN-D-105~\cite{FisherYu2017DilatedRN} for Cityscapes, UperNet101~\cite{TeteXiao2018UnifiedPP} for ADE20K, Unet~\cite{CelebAMask-HQ, ronneberger2015u} for CelebAMask-HQ and DeepLabV2~\cite{LiangChiehChen2014SemanticIS} for COCO-Stuff. 
With the off-the-shelf network, mean Intersection-over-Union (mIoU) is computed based on the generated images and semantic layouts.
The mIoU metric refers to the semantic relevance of the generated images.
However, mIoU highly depends on the capability of the off-the-shelf network.
A strong segmentation network measures the semantic relevance of generated images more correctly.
The reported mIOU is calculated by upsampling the generated images to the same resolution as default input resolution of the off-the-shelf segmentation models, which allows a more reasonable evaluation of the semantic interpretability.

\subsection{Comparison with previous methods} 
We compare our method with several state-of-the-art methods on semantic image synthesis, \emph{i.e.}, SPADE~\cite{TaesungPark2019SemanticIS}, CC-FPSE~\cite{liu2019learning}, INADE~\cite{tan2021diverse} and OASIS~\cite{VadimSushko2020YouON}, \emph{etc.}

\noindent \textbf{Key advantages.} With the help of progressive refinement of the generated results, our methods achieve superior sample quality to previous GAN-based method.
As shown in Table~\ref{tab:comparison}, compared to the most recent methods, our method surpasses them by +2.2, +0.8, +2.0, +1.1 FID score on four datasets, respectively.
Besides the quantitative results, we also conduct the qualitative results on four datasets. We show the results in Figure~\ref{fig:comparison_city}, 
 \ref{fig:comparison_celeba}, \ref{fig:comparison_ade} and \ref{fig:comparison_coco}, we observe that the images generated by our method have better visual performance compared with previous methods.
Under the complex scenes, \emph{i.e.}, fences in front of the building, human faces in the side view and motorcycles with complex structure, our method can generate samples with more reasonable structure and content, which significantly outperforms previous methods.
We also present zoomed-in results of the generated images on Cityscapes dataset.
Notably, our model exhibits more fine-grained details, such as distant cars and traffic lights.

\begin{figure}[t]
    \centering
    \includegraphics[width=\linewidth]{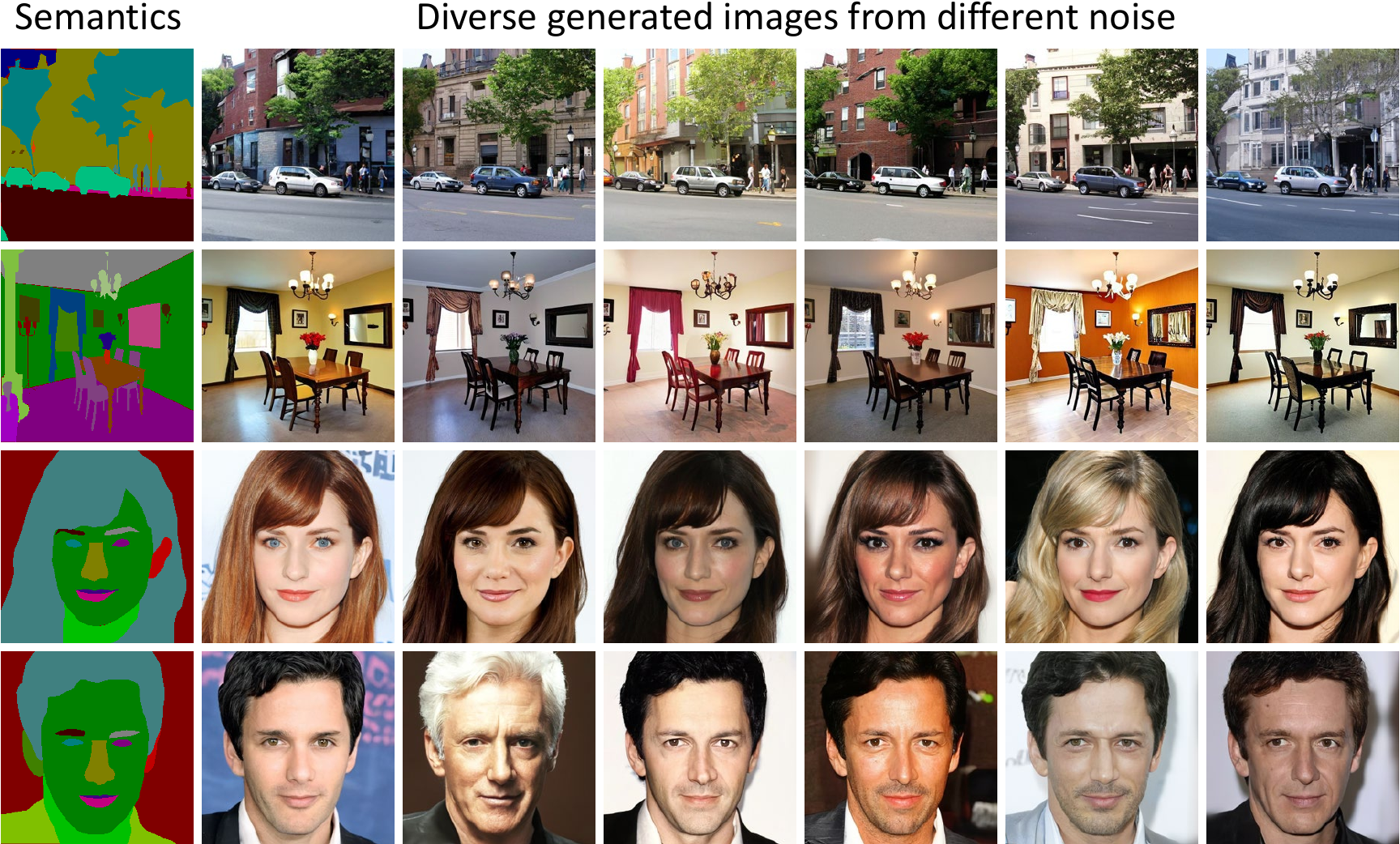}
    \caption{
    Multimodal generation results from our DDPM-based framework.
    it is observed that our method can generate diverse results with high quality.
    }
    \vspace{-4mm}
    \label{fig:diverse}
\end{figure}

\begin{table}[t]
    \footnotesize
    \centering
    \caption{Comparison of our methods with existing solutions in terms of mean Intersection-over-Union (mIoU) on four benchmark datasets.
    The reported mIOU is calculated by upsampling the generated images to the same resolution as default input resolution of the off-the-shelf segmentation models.
    Among these methods, Pix2PixHD and SPADE on the CelebAMask-HQ dataset are based on our implementation, while others are based on released models.
    }
    \vspace{-2mm}
    \begin{tabular}{l @{\hskip 1mm} c  @{\hskip 1mm} c @{\hskip 1mm} c @{\hskip 1mm} c}
    \toprule
    \textbf{Method} & {\textbf{CelebAMask}} & {\textbf{Cityscapes}} & {\textbf{ADE20K}}  & {\textbf{COCO-Stuff}}\\
    \midrule
    {Pix2PixHD~\cite{wang2018high}} & 76.1 & 63.0 & 28.8 & 26.6 \\
    {SPADE~\cite{TaesungPark2019SemanticIS}} & 75.2 & 61.2 & 38.3 & 38.4 \\
    {DAGAN~\cite{tang2020dual}} & 76.6 & 62.4 & 38.1 & n/a\\
    {SCGAN~\cite{wang2021image}} & 75.5 & 55.9 & 41.5 & 44.3 \\ 
    {CLADE~\cite{tan2021efficient}} & 75.4 & 58.6 & 23.9 & 38.8  \\\cdashline{1-5}
    {CC-FPSE~\cite{liu2019learning}} & n/a & 65.2 & 40.6 & 42.9 \\ 
    {GroupDNet~\cite{zhu2020semantically}} & 76.1 & 55.3 & 27.6 & n/a  \\
    {INADE~\cite{tan2021diverse}} & 74.1 & 57.7 & 33.0 & n/a \\
    {OASIS~\cite{VadimSushko2020YouON}} & n/a & 58.3 & \textbf{45.7} & \textbf{46.7}\\     {ControlNet~\cite{zhang2023adding}} & 65.6 & 34.4 & 20.8 & 21.7 \\ \cdashline{1-5}
    {SDM-LoRA~(Ours)} & 64.7 & 48.0 & 16.3 & 12.3 \\
    {SDM~(Ours)} & \textbf{77.0} & \bf{77.5} & 39.2 & 40.2\\
    \bottomrule
    \end{tabular}
    \vspace{-4mm}
    \label{tab:comparison_miou}
\end{table}

We also compare our SDM with recent diffusion-based method, \emph{i.e.}, ControlNet. ControlNet is trained based on pre-trained text-to-image diffusion model, \emph{i.e.}, stable diffusion. Stable diffusion is learned from large-scale dataset LAION, and ControlNet inherits the capability of stable diffusion, which leads to a strong generation capability. We present the quantitative result in Table~\ref{tab:comparison}. It is observed that our SDM achieves better FID and worse LPIPS compared with ControlNet. Qualitative results in Figure~\ref{fig:comparison_city}, \ref{fig:comparison_celeba}, \ref{fig:comparison_ade} and \ref{fig:comparison_coco} also demonstrate that SDM generates more realistic images compared with ControlNet. This is reasonable because ControlNet is designed for various conditional image generation tasks and may not achieve the state-of-the-art performance on the specific conditional generation task, \emph{e.g.}, semantic image synthesis. Besides, with the capabilities of stable diffusion, ControlNet can generate more diverse results.

Furthermore, we conduct a user study to evaluate the visual performance of our method, and three previous methods, \emph{i.e.,} SPADE~\cite{TaesungPark2019SemanticIS}, INADE~\cite{tan2021diverse} and OASIS~\cite{VadimSushko2020YouON}.
There are 20 volunteers participating in this study.
In the study, we present each volunteer 10 pairs of generated results for each pair user study (100 pairs in total) and ask the volunteers to select more high-fidelity results.
The voting results are reported in Table.~\ref{tab:user_study}.
It can be observed that our method is clearly preferred over the competitors in more than 75\% of the time on four benchmark datasets.

\begin{table*}[t]
    \footnotesize
    \centering
    \caption{Ablation studies on the approach to embed the condition information and the classifier-free guidance strategy.
    $\uparrow$ indicates the higher the better, while $\downarrow$ indicates the lower the better.
    }
    \begin{tabular}{c @{\hskip 10mm} c @{\hskip 10mm} c @{\hskip 10mm} c @{\hskip 10mm} c @{\hskip 10mm} c}
    \toprule
    \multicolumn{3}{c}{\textbf{Settings}} & \multicolumn{3}{c}{\textbf{Metrics}} \\
    \cmidrule(r){1-3} \cmidrule(r){4-6}
    {\bf w/ SDM structure} & {\bf Condition position} & {\bf w/Guidance} & {\bf mIoU}$\uparrow$ & {\bf FID}$\downarrow$ & {\bf LPIPS}$\uparrow$ \\
    \midrule
    {$\times$}  & {Dec} & {$\checkmark$} & 23.9 & 39.2 & 0.508 \\
    {$\times$}  & {Enc} & {$\times$} & 34.4 & 30.7 & 0.526 \\
    {$\times$}  & {Enc+Dec} & {$\times$} & 37.4 & 30.1 & 0.522 \\
    {$\checkmark$}  & {Dec} & {$\times$} & 29.7 & 30.8 & \bf{0.593} \\
    {$\checkmark$}  & {Dec} & {$\checkmark$} & \bf{39.2} & \bf{27.5} & 0.524 \\
    \bottomrule
    \end{tabular}
    \label{tab:ablation}
\end{table*}

The other key advantage of our SDM is that our model can inherently generate diverse results given an input semantic mask. We report the results of LPIPS. It is observed that our method surpasses the most diverse methods by +0.035, +0.065, +0.007, +0.190 LPIPS on four datasets, respectively. Furthermore, we present multimodal generation results in Figure~\ref{fig:diverse}, which demonstrates that our method can generate diverse results with high quality.

We also compare our SDM with recent diffusion-based method, \emph{i.e.}, ControlNet. ControlNet is trained based on pre-trained text-to-image diffusion model, \emph{i.e.}, stable diffusion. Stable diffusion is learned from large-scale dataset LAION, and ControlNet inherits the capability of stable diffusion, which lead to a strong generation capability. We present the quantitative result in Table~\ref{tab:comparison}. It is observed that SDM achieve better FID and worse LPIPS compared with ControlNet. Qualitative results in Figure~\ref{fig:comparison_city}, \ref{fig:comparison_celeba}, \ref{fig:comparison_ade} and \ref{fig:comparison_coco} also demonstrate that SDM generates more realistic images compared with ControlNet. This is reasonable because ControlNet is designed for various conditional image generation tasks and may not achieve the state-of-the-art performance on the specific conditional generation task, \emph{e.g.}, semantic image synthesis. And, with the capabilities of stable diffusion, ControlNet can generate more diverse results.

\noindent \textbf{mIoU comparison and analysis.}
To evaluate the semantic correspondence, we utilize the mean Intersection-over-Union (mIoU) with the off-the-shelf segmentation models.
On the CelebAMask-HQ and Cityscapes dataset, we achieve 77.0 and 77.5 mIoU, surpassing previous sota by \textbf{+0.4} and \textbf{+12.3}, indicating the superior performance of our methods for generating highly semantic correlated images.
On the ADE20K and COCO-Stuff dataset, our method shows a weaker performance on mIoU compared with some existing methods. However, we observe that our generated images in Figure~\ref{fig:comparison_ade} and~\ref{fig:comparison_coco} have a strong semantic correlation with input mask visually, which are at least comparable with those of previous methods. One possible explanation is that the semantic segmentation model used for evaluating is not that strong, we checked that the segmentation model and find that this model only achieves $35.3$ and $39.0$ mIoU on the ground-truth images, which is much lower than the model we used on CelebAMask-HQ and Cityscapes dataset. 
We will show randomly selected 100 results in the supplementary material to further verify this issue.

\subsection{Ablation Studies}
\label{sec:ablation}
We perform ablative experiments to verify the effectiveness of several important designs in our framework, \emph{i.e.}, the approach to embed the condition information, the position to embed the condition information and the classifier-free guidance strategy.
We report the qualitative and quantitative results on the ADE20K dataset.

\begin{figure}[t]
    \centering
    \includegraphics[width=\linewidth]{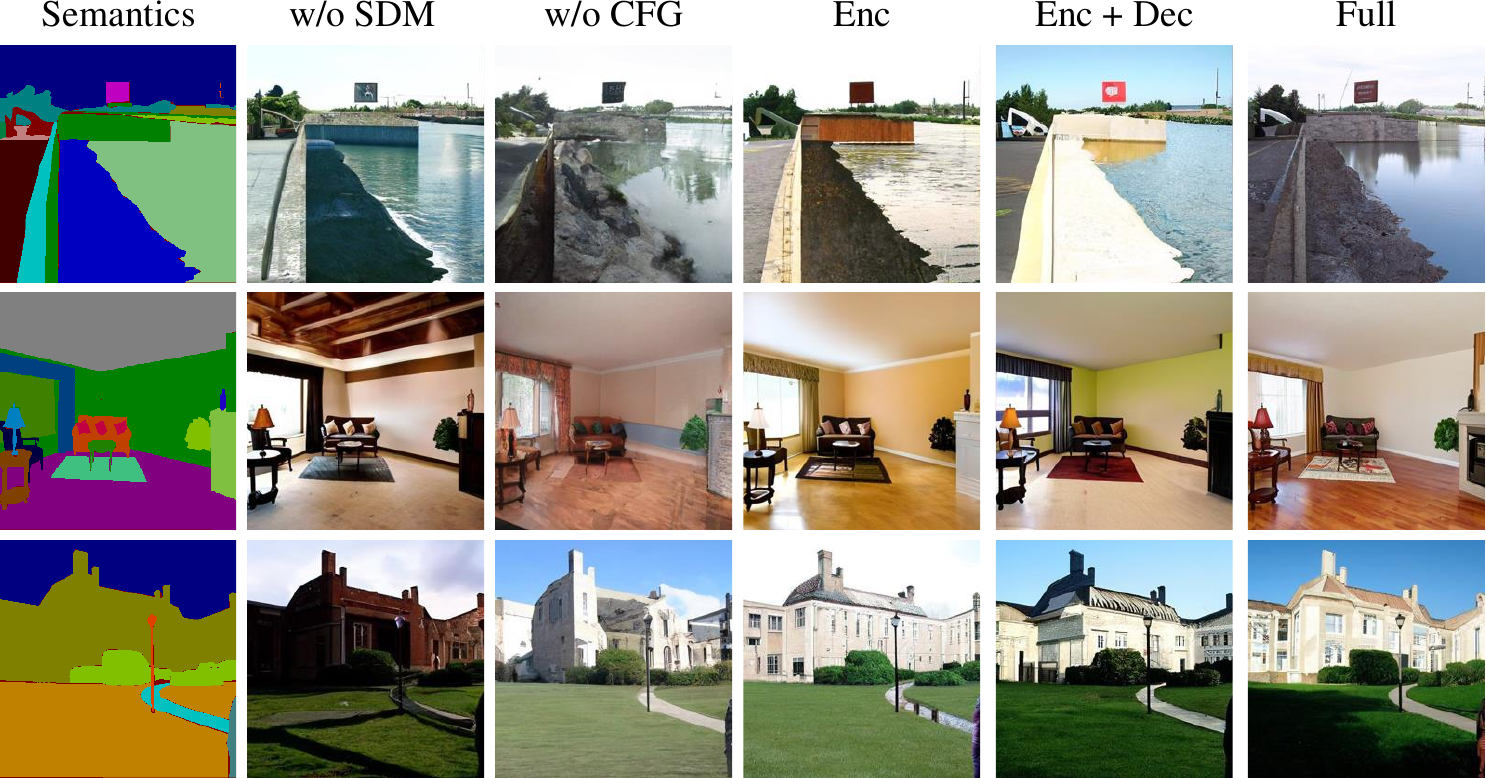}
    \caption{
    Qualitative results on ablative experiments.
    The images generated under full settings show superior performance compared with those w/o SDM structure or classifier-free guidance and those injecting conditions at the encoder or at both the encoder and decoder.
    }
    \vspace{-4mm}
    \label{fig:ablation}
\end{figure}

\noindent\textbf{Condition Embedding.} To evaluate the importance of embedding the condition information independent of the noisy image, we design a baseline variant as comparison.
As an alternative, we directly apply the conditional DDPM in \cite{saharia2021palette,saharia2021image}, which directly concatenates the semantic label map with the noisy image as input.
The quantitative results are reported in Table~\ref{tab:ablation}.
It is observed that our semantic diffusion model highly outperforms previous conditional DDPM on all the metrics.
Additionally, we analyze the visual results between these two variants.
In Figure~\ref{fig:ablation}, it can be seen, by embedding the semantic label map in a multi-layer spatially-adaptive manner, the generated image exhibit superior visual quality on fidelity and correspondence with the semantic label map.

\noindent\textbf{Embedding Position.} To uncover the position to effectively embed the condition information, we design three variants which embed the condition information at the encoder, at the decoder and at both the encoder and decoder, respectively. The quantitative results are reported in Table~\ref{tab:ablation}. It can be seen that embedding the condition information at the decoder part achieve the superior performance. We present the qualitative comparison in Figure~\ref{fig:ablation}, which also demonstrates that the variant which embeds the condition at the decoder generates images with better fidelity and semantic correspondence.

\begin{figure}[t]
    \centering
    \includegraphics[width=\linewidth]{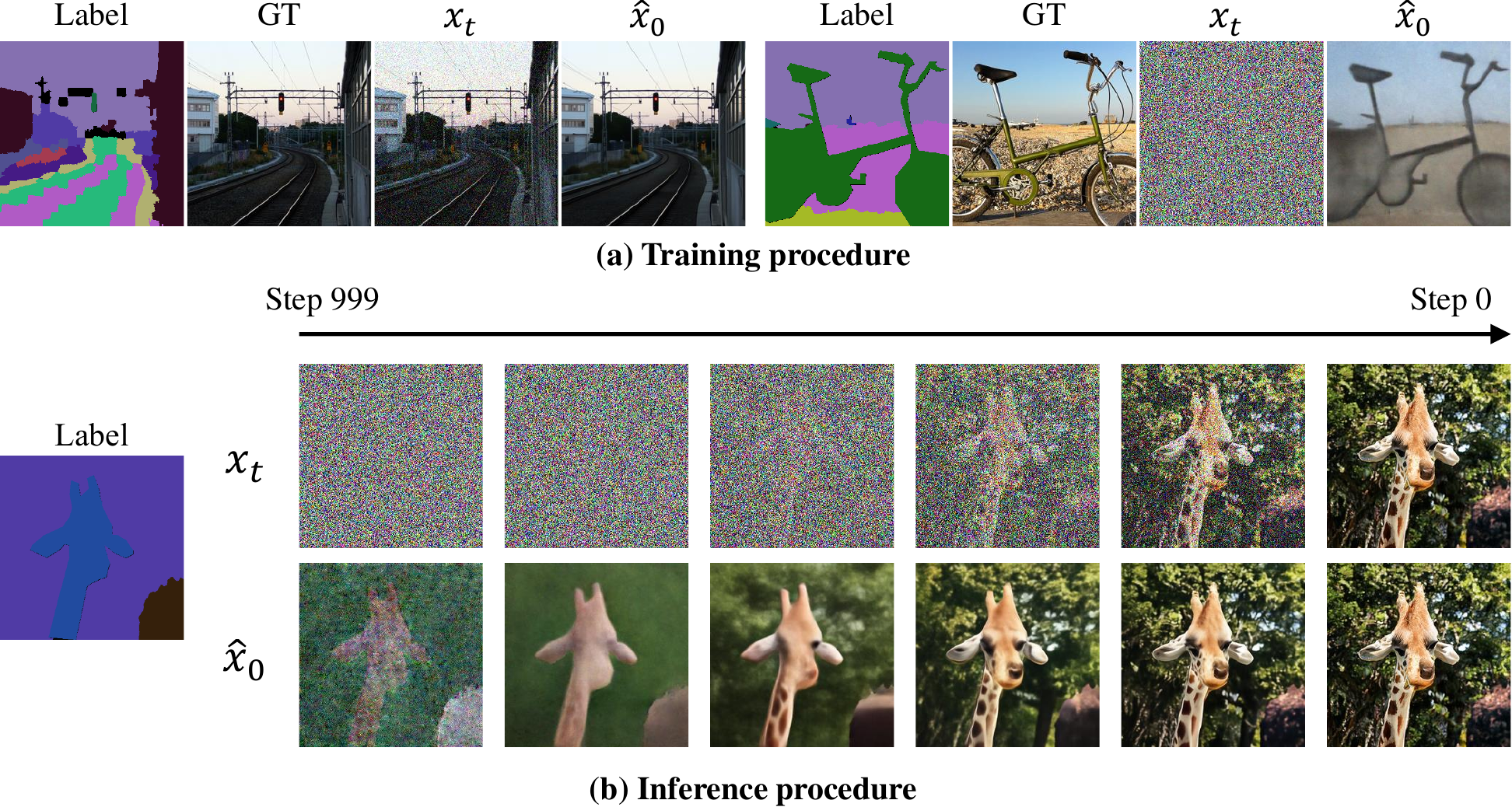}
    \caption{
    Visualization of the intermediate results in both training and diffusion procedure.
    During training, SDM learns to produce noise-free images by predicting the noise involved.
    In the diffusion process, SDM generates realistic images with iterative refinement.
    }
    \vspace{-3mm}
    \label{fig:intermediate}
\end{figure}

\noindent\textbf{Importance of classifier-free guidance.} Furthermore, we study the effectiveness of the classifier-free guidance strategy.
We take the variant without classifier-free guidance as a comparison.
From Table~\ref{tab:ablation}, the classifier-free guidance highly improves the mIoU and FID metrics at the expense of little LPIPS. 
In Figure~\ref{fig:ablation}, we present the qualitative results on the classifier-free guidance strategy.
The images generated with classifier-free guidance better exhibit semantic information and generates more structured content. 
This further improves the visual effects of generated images compared with those without classifier-free guidance.

\subsection{LoRA-like SDM}
To verify the generalization of our proposed SDM, we introduce a LoRA-like semantic diffusion model named SDM-LoRA based on similar design of the proposed SDDResblock to finetune the pre-trained text-to-image model, \emph{i.e.}, stable diffusion. In the design of SDM-LoRA, we add LoRA module to stable diffusion and modify the group normalization to inject semantic layout in a spatial-adaptive manner. We present the quantitative results in Table~\ref{tab:comparison} and \ref{tab:comparison_miou}. It is observed that SDM-LoRA also achieves encouraging performance, which further demonstrates the generalization of our approach.

\subsection{Visualization of SDM}
For better understanding the entire process of SDM, we visualize the intermediate results in both training and diffusion procedures. 
As shown in Figure~\ref{fig:intermediate}, SDM learns to produce noise-free images by predicting the noise involved in training and generate realistic images with iterative refinement.
During training procedure, in small timestep, with a small magnitude of noise involved, SDM learns to recover the details of the image.
In large timestep, SDM restores the coarse-grained shape and outline of the image due to the difficulty of generating from the highly noisy image.
During diffusion procedure, SDM first produces the rough outline of the image in the early stage of diffusion, and then iteratively refines the details of the generated image.

\begin{figure}[t]
    \centering
    \includegraphics[width=\linewidth]{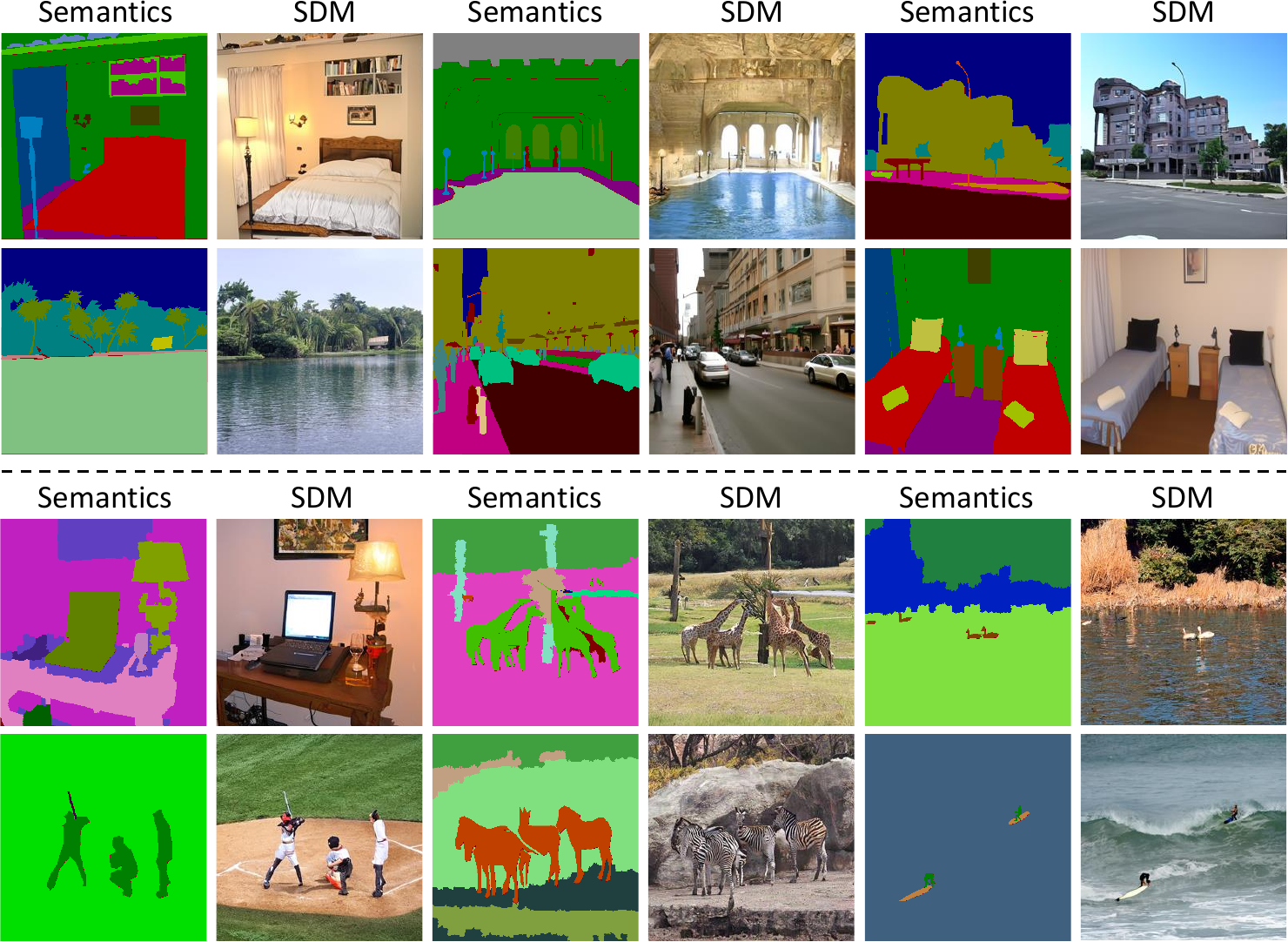}
    \caption{Generation results based on DPM-Solver~\cite{lu2022dpm} on ADE20K and COCO-Stuff.
    SDM is able to produce realistic images with fewer sampling steps, \emph{i.e.}, 100 steps.
    }
    \vspace{-3mm}
    \label{fig:dpms}
\end{figure}

\begin{table}[t]
    \footnotesize
    \centering
    \caption{\textbf{Quantitative results with standard DDPM sampler ($T=1000$) and DPM-Solver ($T=100$) on semantic image synthesis.}
    $\uparrow$ indicates the higher the better, while $\downarrow$ indicates the lower the better.
    }
    \resizebox{\linewidth}{!}{
    \begin{tabular}{l cc cc}
    \toprule
    \multirow{2}{*}{\textbf{Method}} &  \multicolumn{2}{c}{\textbf{ADE20K}} & \multicolumn{2}{c}{\textbf{COCO-Stuff}} \\
    \cmidrule(r){2-3} \cmidrule(r){4-5}  
    & {\bf FID}$\downarrow$ & {\bf mIOU}$\uparrow$ & {\bf FID}$\downarrow$ & {\bf mIOU}$\uparrow$ \\
    \midrule
    {DDPM~($T=1000$)} & 27.5 & 39.2 & 15.9 & 40.2  \\
    {DPMSolver~($T=100$)} & 46.8 & 35.4 & 22.5 & 32.2  \\
    \bottomrule
    \end{tabular}
    }
    \label{tab:fast_sample}
\end{table}

\subsection{Fast Sampling}
Diffusion models generates both high-fidelity and diverse images at the expense of sampling speed compared with GAN-based methods.
To tackle the inherently slow sampling rate of the diffusion models, some fast sampling strategies, \emph{e.g.}, DDIM~\cite{song2020denoising} and DPM-solver~\cite{lu2022dpm} are proposed based on diffusion ordinary differential equations (ODEs).
These strategies can also directly applied to pretrained SDM for more effecient sampling.
We present quantitative and qualitative results by DPM-solver~(T=100) in Table~\ref{tab:fast_sample} and Figure~\ref{fig:dpms}, respectively.
From the figure, it is observed that SDM is able to generate realistic images based on fewer steps.
Based on DPM-solver, SDM obtains a sampling speed of 17.5s/image on single V100 GPU.

\begin{figure}[t]
    \centering
    \includegraphics[width=\linewidth]{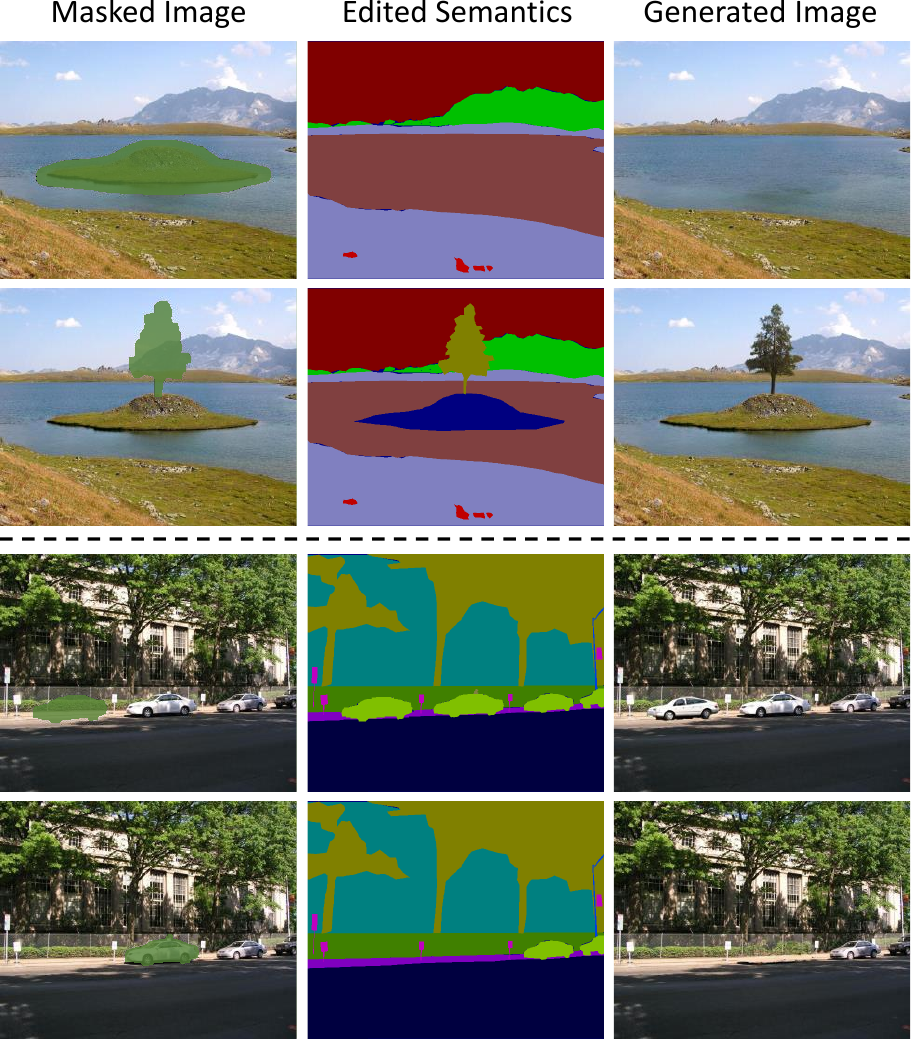}
    \vspace{-6mm}
    \caption{
    Semantic image editing examples from SDM. 
    The green region is erased, and the model inpaints it based on the edited semantic map. 
    Our model produce a realistic completion matching both the semantic label map and surrounding context.
    }
    \label{fig:editing}
    \vspace{-4mm}
\end{figure}

\subsection{Controlled Generation}
We study our trained semantic diffusion model on the controlled image generation, \emph{i.e.}, semantic image editing.
Considering a real image with corresponding semantic label map, we add or remove objects by modifying the semantic label map.
To ensure the harmony between the inpainted area and the original image, inspired by RePaint~\cite{lugmayr2022repaint}, we inject the unmasked image into the diffusion process using a resampling operation as follows,
\begin{equation}
    \hat{y}_t = m \cdot y_t + (1 - m) \cdot y_t^o,
\end{equation}
where $y_t$ and $\hat{y}_t$ are the original and resampled noisy image, respectively. $y_t^o$ is the noisy version of the image before editing and $m$ is a mask which denotes the inpainted area. We adopt the resampled noisy image $\hat{y}_t$ for further diffusion process and compute the noisy image at timestep $t - 1$, $y_{t-1}$.

According to aforementioned approach, our trained model inpaints the area conditioned on the edited semantic label map, which enables image manipulation of real images through user interaction.
We show the semantic image editing results in Figure~\ref{fig:editing}.
It is observed that our model can produce a realistic completion matching both the semantic label map and surrounding context.

\section{Limitation and Future work}
Our semantic diffusion model is based on pixel-level diffusion model trained from scratch on the datasets, which makes our approach has good performance on the dataset but not general on natural images. Current large-scale pre-trained text-to-image generation models have strong inherent generation capability and semantic understanding. Therefore, leveraging inherent capabilities of large-scale pre-trained models may be the direction of future exploration on semantic image synthesis. We attempt a LoRA-like semantic diffusion model, \emph{i.e.}, SDM-LoRA, based on pre-trained stable diffusion in the \emph{Experiment} Section and will make further attempts in the future.

Furthermore, we propose to process semantic layout and
noisy image differently, by multi-layer spatially-adaptive normalization operators. In our framework, we implement spatially-adaptive injection via SPADE, which was originally proposed for GANs. In GAN-based methods, many new spatially-adaptive injection methods have been proposed as improved variants of SPADE, \emph{e.g.}, INADE, CLADE and CC-FPSE. These variants have been proven to be effective on GANs and future research may focus on combining these improved variants to improve the performance of SDM.

\section{Conclusions}
In this paper, we present the first attempt to explore diffusion model for the problem of semantic image synthesis and design a novel framework named Semantic Diffusion Model (SDM).
Specifically, we propose a new network structure to handle noisy input and semantic mask separately and precisely to fully leverage the semantic information.
Furthermore, we introduce classifier-free guidance during sampling process, significantly improve the generation quality and semantic interpretability in semantic image synthesis.
Extensive experiments on four benchmark datasets demonstrate the effectiveness of our method.
Our method achieves state-of-the-art performance in terms of FID and LPIPS metrics and shows better visual quality of generated images compared with previous methods.

\bibliographystyle{IEEEtran}
\bibliography{reference}

%

\begin{IEEEbiography}[{\includegraphics[width=1in,height=1.25in,clip,keepaspectratio]{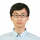}}]{Wengang Zhou}
received the B.E. degree in electronic information engineering from Wuhan University, China, in 2006, and the Ph.D. degree in electronic engineering and information science from University of Science and Technology of China (USTC), China, in 2011. 
From September 2011 to 2013, he worked as a post-doc researcher in Computer Science Department at the University of Texas at San Antonio. He is currently a Professor at the EEIS Department, USTC.

His research interests include multimedia information retrieval and computer vision.
\end{IEEEbiography}

\begin{IEEEbiography}
[{\includegraphics[width=1in,height=1.25in,clip,keepaspectratio]{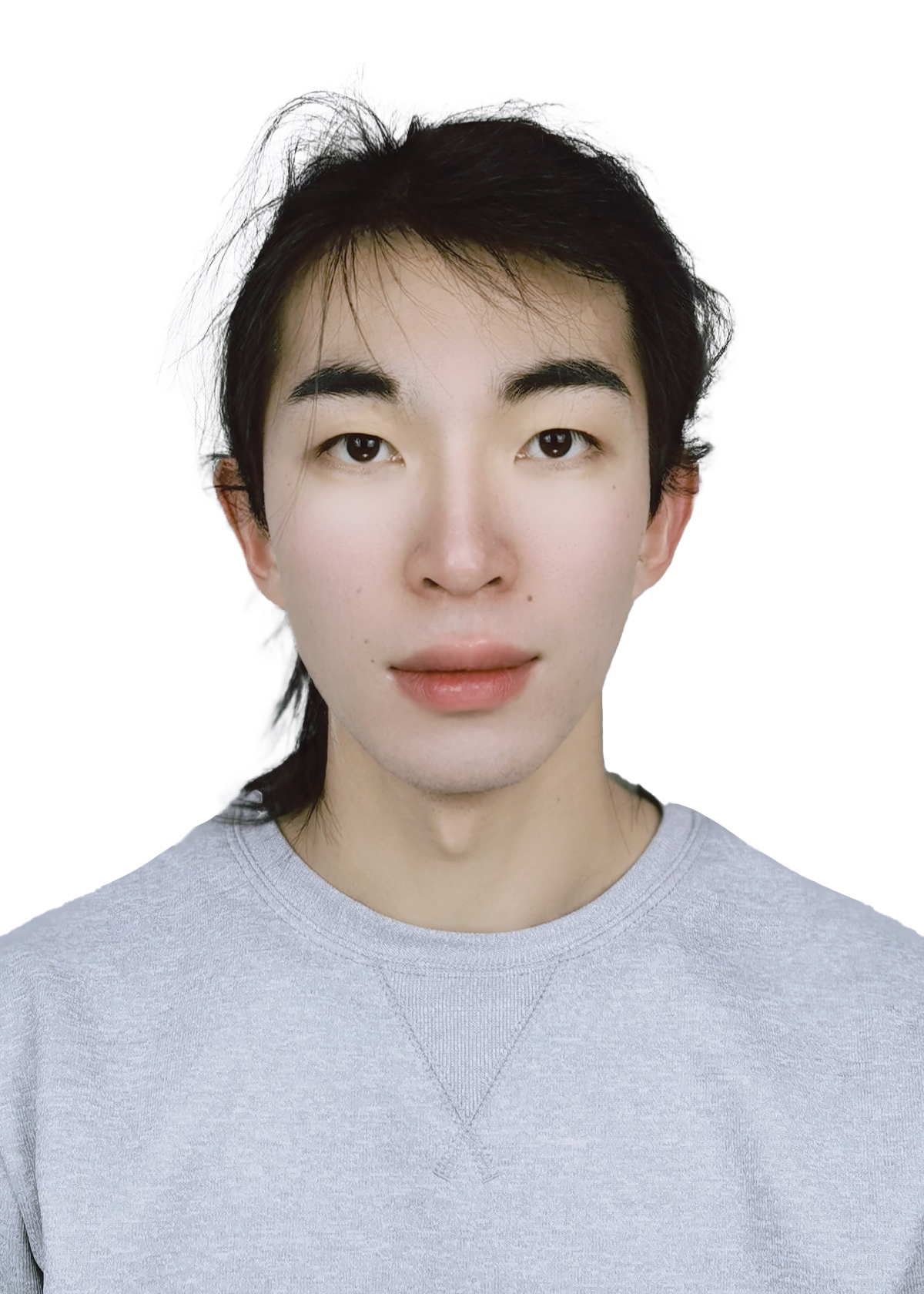}}]{Weilun Wang}
received the B.Eng. and Ph.D. degrees in electronic engineering from the University of Science and Technology of China, Hefei, China, in 2019, and 2024. He is currently a research and development engineer with ByteDance Inc.

His research interests include denoising diffusion probabilistic models, generative adversarial nets, image generation, pose-guided person generation and image-to-image translation.
\end{IEEEbiography}


\begin{IEEEbiography}[{\includegraphics[width=1in,height=1.25in,clip,keepaspectratio]{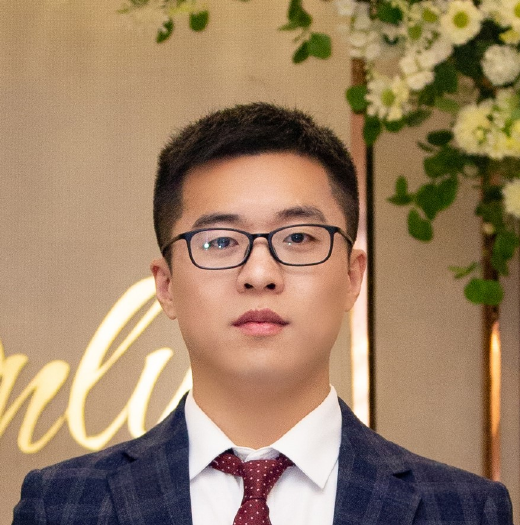}}]{Jianmin Bao}
received the B.Eng. and Ph.D. degrees in electronic engineering from the University of Science and Technology of China, Hefei, China, in 2014, and 2019, respectively. 
He is currently a Senior Researcher with Microsoft Research Asia. 

His research interests mainly include generative adversarial nets, image generation, low-level image processing, face related tasks, and general representation learning.
\end{IEEEbiography}

\begin{IEEEbiography}
[{\includegraphics[width=1in,height=1.25in,clip,keepaspectratio]{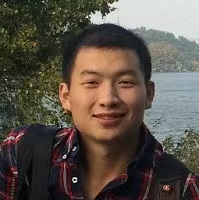}}]{Dongdong Chen} is a Principal Research Manager from Microsoft GenAI. He received his Ph.D. degree under the joint phd program between University of Science and Technology of China and MSRA. His research interests mainly include generative mode, large-scale pretraining, and general representation learning. He is the Associate Editor of Pattern Recognition and IEEE Transactions on Multimedia.
\end{IEEEbiography}

\begin{IEEEbiography}
[{\includegraphics[width=1in,height=1.25in,clip,keepaspectratio]{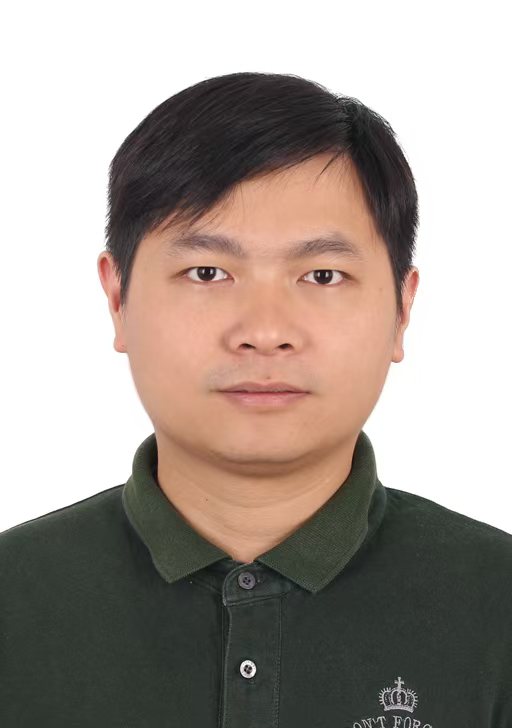}}]{Dong Chen} is currently the Principal Researcher Manager of the Visual Computing Group at Microsoft Research Asia. He received the B.S. and Ph.D. degree from the University of Science and Technology of China in 2010, 2015. He joined Microsoft Research in July 2015. His team is engaged in research on image synthesis models such as Generative Adversarial Networks (GAN), Denoising Diffusion Probabilistic Model (DDPM), and Generative Artificial Intelligence (AIGC). He has published more than 50 papers in international conferences such as CVPR/ICCV/ECCV. Holds 8 patents. Multiple research results have been used in products such as Microsoft Cognitive Services, Windows Hello face unlock in Windows 10, and Microsoft Designer.
\end{IEEEbiography}

\begin{IEEEbiography}
[{\includegraphics[width=1in,height=1.25in,clip,keepaspectratio]{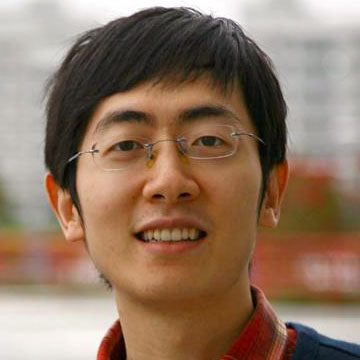}}]{Lu Yuan} received his PhD degree from the Department of Computer Science and Engineering at the Hong Kong University of Science and Technology in 2009. Before that, he received his MS degree at TsingHua University. He wa a Principal Research Manager in Microsoft Redmond. Currently, he is a research scientist in Meta. His research interests include computer vision, applied machine learning and computational photography.
\end{IEEEbiography}

\begin{IEEEbiography}[{\includegraphics[width=1in,height=1.25in,clip,keepaspectratio]{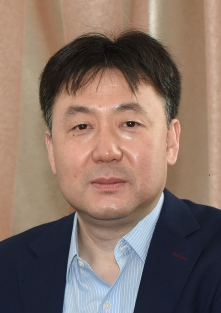}}]{Houqiang Li}
received the B.S., M.Eng., and Ph.D. degrees in electronic engineering from the University of Science and Technology of China, Hefei, China, in 1992, 1997, and 2000, respectively, where
he is currently a Professor with the Department of Electronic Engineering and Information Science.

His research interests include multimedia search, image/video analysis, video coding and communication. 
He has authored and co-authored over 200 papers in journals and conferences. 
He is the winner of National Science Funds (NSFC) for Distinguished Young Scientists, the Distinguished Professor of Changjiang Scholars Program of China, and the Leading Scientist of Ten Thousand Talent Program of China. 
He served as an Associate Editor of the IEEE TRANSACTIONS ON CIRCUITS AND SYSTEMS FOR VIDEO TECHNOLOGY from 2010 to 2013. 
He served as the TPC Co-Chair of VCIP 2010, and he will serve as the General Co-Chair of ICME 2021. 
He is the recipient of National Technological Invention Award of China (second class) in 2019 and the recipient of National Natural Science Award of China (second class) in 2015. 
He was the recipient of the Best Paper Award for VCIP 2012, the recipient of the Best Paper Award for ICIMCS 2012, and the recipient of the Best Paper Award for ACM MUM in 2011.
\end{IEEEbiography}




\end{document}